\documentclass[10pt,twocolumn,letterpaper]{article}

\usepackage[pagenumbers]{cvpr} 
\usepackage{algorithm}
\usepackage[noend]{algpseudocode}
\usepackage{xr}
\usepackage{colortbl}
\usepackage{xcolor}
\usepackage{multirow}
\usepackage{multicol}
\definecolor{lightblue}{RGB}{226,240,248}
\definecolor{lightgray}{gray}{0.85}
\newcommand{\name}{\textit{AutoRefiner}\xspace}
\definecolor{mediumslateblue}{rgb}{0.43, 0.43, 1}
\definecolor{salmon}{rgb}{1, 0.42, 0.42}
\newcommand{\best}[1]{\textbf{\textcolor{salmon}{#1}}}
\newcommand{\secondbest}[1]{\textcolor{mediumslateblue}{\textbf{#1}}}
\usepackage{makecell}

\makeatletter
\DeclareRobustCommand\onedot{\futurelet\@let@token\@onedot}
\def\@onedot{\ifx\@let@token.\else.\null\fi\xspace}

\def\eg{e.g\onedot} 
\def\ie{i.e\onedot}

\makeatother
\usepackage{csquotes}

\newcommand{\bepsilon}{{\mathbf{\epsilon}}}

\newcommand{\bv}{{\mathbf{v}}}

\newcommand{\bx}{{\mathbf{x}}}

\newcommand{\bI}{{\mathbf{I}}}

\newcommand{\bX}{{\mathbf{X}}}

\newcommand{\cN}{\mathcal{N}}

\usepackage{pifont}
\newcommand{\cmark}{\textcolor{ForestGreen}{\ding{51}}} 
\newcommand{\xmark}{\textcolor{BrickRed}{\ding{55}}}    

\newcommand{\newparagraph}[1]{\noindent \textbf{#1}}
\definecolor{cvprblue}{rgb}{0.21,0.49,0.74}
\usepackage[pagebackref,breaklinks,colorlinks,allcolors=cvprblue]{hyperref}
\makeatletter
\let\orig@maketitle\@maketitle
\def\@maketitle{%
  \orig@maketitle%
  \begingroup
    \renewcommand\thefootnote{\fnsymbol{footnote}}%
    \@thanks
  \endgroup
}
\makeatother

\def\paperID{} 
\def\confName{CVPR}
\def\confYear{2026}

\title{AutoRefiner: Improving Autoregressive Video Diffusion Models via \\ Reflective Refinement over the Stochastic Sampling Path}

\author{$\text{Zhengyang Yu}^{\dagger,*}$ \quad $\text{Akio Hayakawa}^{\ddagger}$ \quad $\text{Masato Ishii}^{\ddagger}$ \quad $\text{Qingtao Yu}^{\dagger}$\\
$\text{Takashi Shibuya}^{\ddagger}$\quad $\text{Jing Zhang}^\dagger$, $\text{Yuki Mitsufuji}^{\ddagger,\S}$\\
$^{\dagger}$ Australian National University \quad  $\ddagger$Sony AI \quad  $\S$Sony Group Corporation\\
{\tt\small $\{$zhengyang.yu,terry.yu,jing.zhang$\}$@anu.edu.au}\\
{\tt\small $\{$akio.hayakawa,masato.a.ishii,takashi.tak.shibuya,yuhki.mitsufuji$\}$@sony.com}
}
\newcommand\blfootnote[1]{%
  \begingroup
  \renewcommand\thefootnote{}\footnote{#1}%
  \addtocounter{footnote}{-1}%
  \endgroup
}

\usepackage[accsupp]{axessibility}  
\begin{document}
\twocolumn[{
    \begin{center}
    \maketitle
        \vspace{-8mm}
            \includegraphics[width=\linewidth]{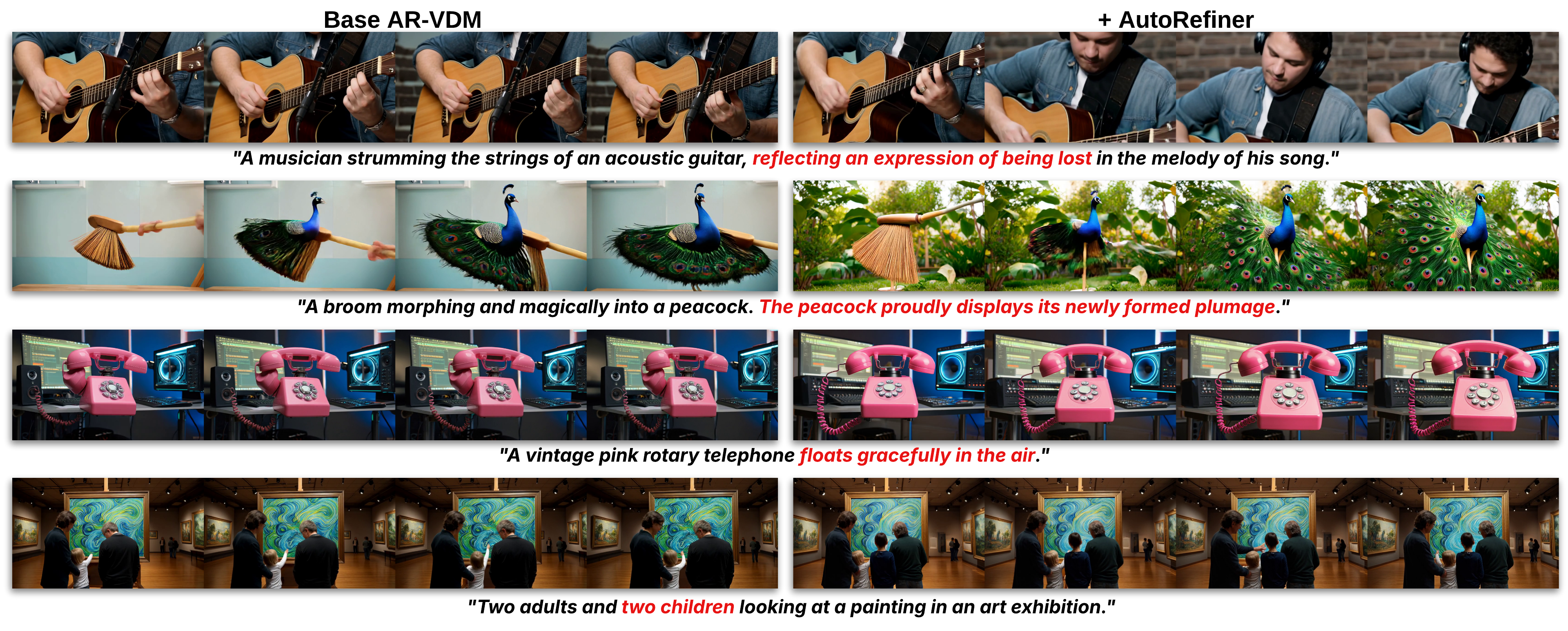}
            \captionof{figure}{\name efficiently enhances base AR-VDMs through feedforward refinement of noises sampled along the intermediate denoising path, achieving improved video fidelity across aspects such as perceptual quality and scene–motion alignment with text input.}
            \label{fig:1}
    \end{center}
}]

\blfootnote{\thanks{${}^*$Work done during an internship at Sony AI}}

\begin{abstract}
Autoregressive video diffusion models (AR-VDMs) show strong promise as scalable alternatives to bidirectional VDMs, enabling real-time and interactive applications. Yet there remains room for improvement in their sample fidelity. A promising solution is inference-time alignment, which optimizes the noise space to improve sample fidelity without updating model parameters. Yet, optimization- or search-based methods are computationally impractical for AR-VDMs. Recent text-to-image (T2I) works address this via feedforward noise refiners that modulate sampled noises in a single forward pass. \textbf{Can such noise refiners be extended to AR-VDMs?} We identify the failure of naïvely extending T2I noise refiners to AR-VDMs and propose \name---a noise refiner tailored for AR-VDMs, with two key designs: pathwise noise refinement and a reflective KV-cache. Experiments demonstrate that \name serves as an efficient plug-in for AR-VDMs, effectively enhancing sample fidelity by refining noise along stochastic denoising paths.
\end{abstract}    
\section{Introduction}
\label{sec:intro}
Recent advances in video diffusion models (VDMs) enabled photorealistic and temporally coherent video synthesis~\cite{ho2022video,hunyuan,chen2023videocrafter1,chen2024videocrafter2}. However, most VDMs employ bidirectional temporal attention, generating all frames jointly without respecting causal order, which limits their use in real-time or interactive applications. Autoregressive video diffusion models (AR-VDMs)~\cite{diffusionforcing,causvid,selfforcing} have emerged as a promising alternative, which condition on previously generated frames to predict future ones, forming the foundation for applications such as neural game engines~\cite{oasis2024,cheng2025playing}, cinematic creation~\cite{Captaincinema,moviedreamer}, 
and world simulation~\cite{po2025long,xiao2025worldmem,po2025long}. Yet, there remains substantial room to improve the sample fidelity of AR-VDMs, \eg, maintaining high perceptual fidelity, realistic motion dynamics, and semantic alignment.  
While scaling model capacity and data have been the dominant route to improving sample fidelity of diffusion models, a rapidly growing orthogonal direction focuses on improving generation capacity at inference time. This line of work, known as \textit{inference-time alignment}~\cite{uehara2025inference}, enhances sample fidelity or optimizes specific reward functions at inference time without fine-tuning the base model. The core idea is to exploit the stochasticity inherent in the noise sampling process of diffusion models. Existing methods identify ``golden noises" either by optimizing noise as a learnable latent via reward gradients~\cite{reno,D-Flow,titan}, or by searching for noise samples that yield higher reward scores~\cite{qi2024not,bon,karthik2023if,guo2025trainingfreeguidancedifferentiabilityscalable}. While both strategies have proven effective for text-to-image (T2I) models, applying them to AR-VDMs is computationally impractical: optimization-based methods require costly gradient propagation through multiple denoising steps---easily leading to out-of-memory issues for large VDMs---whereas search-based methods demand over hundreds of trials per generation, with computational cost further scaling with the temporal rollout length of AR-VDMs. 

We therefore consider a more practical objective---learning a feedforward network for AR-VDMs that directly maps a random noise to its locally optimized counterpart in a single forward pass, which we call a \textbf{noise refiner}, achieving a similar effect to inference-time alignment with only minor degradation in inference efficiency. While the similar idea of a noise refiner has been explored in T2I works~\cite{goldennoise,hypernoise,ahn2024noise}, we observe that their naïve extension to AR-VDMs yields only moderate or worse performance and reward hacking issues. We attribute the culprit to refining the \textit{initial} noise, a design suited to deterministic samplers. Such a strategy fails for step-distilled AR-VDMs that rely on stochastic sampling to maintain sample fidelity (\cref{fig:ode_vs_stoch}), where the intermediate stochasticity dilutes the influence of the initial noise. We additionally observe that initial noise refinement is prone to reward hacking issues such as motion degradation and `grid-like' artifacts (\cref{fig:reward_comparison}). 

To address these limitations, we introduce \name---the first noise refinement framework tailored for AR-VDMs. The failure of initial noise refinement and the critical role of intermediate noises in the sampling process with AR-VDMs motivate our design of \textbf{pathwise noise refinement}, which refines the noises sampled along the intermediate denoising path rather than the initial noise. Built upon the base AR-VDM and trained via efficient LoRA~\cite{hu2021lora} fine-tuning, \name naturally inherits the base model’s autoregressive architecture, enabling it to leverage the KV cache for temporal conditioning. We extend this capability with a \textbf{reflective KV-cache} mechanism, allowing the refiner to condition on both historical frames and the previous denoising state. This design equips \name with a context-aware and self-reflection ability to identify locally optimized noises more effectively. Our contributions are:
\begin{enumerate}
    \item We identify and analyze the failure of T2I noise refiners when extended to AR-VDMs. 
    \item We propose \name, a noise refinement framework for AR-VDMs featuring pathwise noise refinement and a reflective KV cache mechanism that jointly enables context-aware noise refinement along the denoising path. 
    \item We demonstrate that \name serves as a plug-in that improves the performance of two mainstream AR-VDM paradigms---Diffusion-Forcing~\cite{diffusionforcing} and Self-Forcing~\cite{selfforcing} without incurring costly inference.
\end{enumerate}
\section{Related Work}
\label{sec:related_work}
\newparagraph{Bidirectional VDMs.}
Building on the success of image diffusion models~\cite{imagen,glide,dalle,stablediffusion,sd3,flux}, video diffusion models (VDMs) generalize diffusion models to synthesize temporally coherent and high-fidelity videos. These include pixel-space models~\cite{ho2022video,ho2022imagen,singer2022make,zhang2023show} and latent-space models~\cite{he2022latent,guo2023animatediff,wang2023modelscopetexttovideotechnicalreport,wang2023lavie,chen2023videocrafter1,chen2024videocrafter2,Open-Sora,mochi,hunyuan,hong2022cogvideo,yang2024cogvideox,wan2025wan}.

\newparagraph{AR VDMs.} The temporal continuity of videos naturally lends itself to a next-frame prediction scheme for video generation. Rolling Diffusion~\cite{rollingdiffusion} and its variants~\cite{rollingforcing,fifo,kodaira2025streamdit,pavdm,Ardiffusion,framepack} adopt a sliding-window denoising strategy. However, these methods depend on overlapping windows to mitigate error accumulation, leading to substantial latency. Inspired by the success of LLMs~\cite{brown2020language}, recent works explore hybrid diffusion–AR training schemes. Diffusion Forcing (DF)~\cite{diffusionforcing} and its extensions~\cite{song2025history,causvid,skyreels,magi} train VDMs with independently sampled per-frame noise levels, enabling next-frame denoising during inference conditioned on clean past frames. CausVid~\cite{causvid} employs DF to train causal few-step VDMs via step-distillation~\cite{dmd} from a bidirectional teacher. Self-Forcing (SF)~\cite{selfforcing} and its variants~\cite{rollingforcing,longlive,selfforcing++} address the train–test discrepancy of DF by simulating rollouts during training, improving extrapolative consistency and alleviating exposure bias~\cite{schmidt2019generalization}.
\newparagraph{Diffusion Inference-Time Alignment.}
Similar to the inference-time scaling behavior observed in LLMs~\cite{brown2024large,snell2024scaling}, recent studies~\cite{bon,oshima2025inference,scalingnoise,karthik2023if,qi2024not} have shown that scaling compute to search for ``golden noises" during inference can substantially enhance diffusion sampling quality. Another line of work~\cite{titan,reno} iteratively optimizes noise with differentiable reward models~\cite{pickscore,imagereward,hpsv2,internvid}, improving human preference alignment or controllability. While these works achieve notable quality gains, they often incur substantial computational overhead. Recent efforts have proposed training noise refiners~\cite{ahn2024noise,goldennoise,hypernoise} that directly predict optimized input noise in a single forward pass, removing the need for costly search or optimization at inference time. 
\begin{figure*}[!ht]
    \centering
        \includegraphics[width=\linewidth]{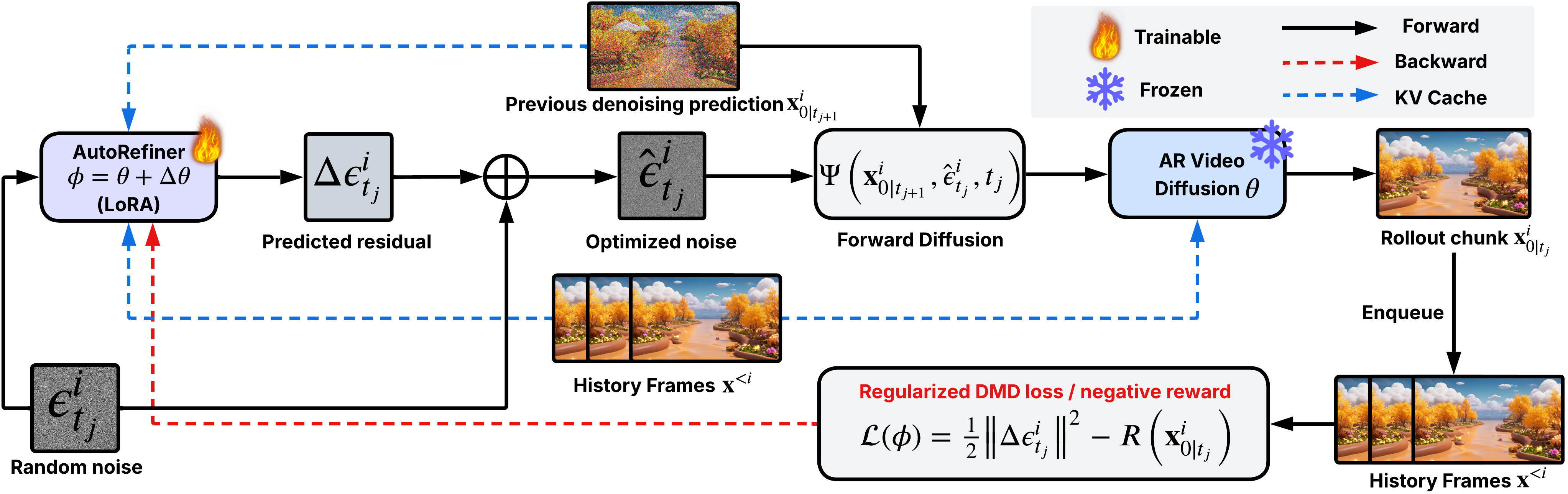}
        \vspace{-6mm}
        \caption{\textbf{The \name method.} \name is a feed-forward plug-in network for enhancing pretrained AR-VDMs. Without modifying the base model parameters, it is trained to maximize distribution-matching or reward-based video-fidelity scores by refining randomly sampled noises along the intermediate denoising trajectory (\cref{sec:traj_noise_refine}). Built through LoRA fine-tuning, \name inherits the autoregressive structure of the base AR-VDM and extends its KV cache mechanism to a reflective KV cache (\cref{fig:ref_kv}), allowing it to condition on both historical frames and the previous denoising prediction for more effective, context-aware noise refinement (\cref{sec:ref_kv}).}
        \label{fig:overview}
\end{figure*}
\section{Method}
\label{sec:method}
\label{sec:preliminaries}
\subsection{Preliminaries: Autoregressive VDMs}
\label{sec:preliminaries}

\newparagraph{Next-chunk Prediction by Autoregression.} An autoregressive video diffusion model is a hybrid framework that approximates the joint distribution of temporally ordered video frames or chunks. Given a video sequence of length $N$, denoted as $\bx^{1: N}=\left(\bx^1, \bx^2, \ldots, \bx^N\right)$, the joint distribution can be factorized into a product of conditionals $p\left(\bx^{1: N}\right)=\prod_{i=1}^N p\left(\bx^i \mid \bx^{<i}\right)$. Each conditional $p\left(\bx^i \mid \bx^{<i}\right)$ is approximated by a denoising model $G_\theta$ 
\begin{equation}
\bx^i=f_{\theta, t_1} \circ f_{\theta, t_2} \circ \ldots \circ f_{\theta, t_T}\left(\bx_{t_T}^i\right)\sim q_\theta\left(\bx^i \mid \bx^{<i}\right),
\label{eq:conditional}
\end{equation}
where $\bx_{t_T}^i\sim\mathcal{N}\left(0, I\right)$ is the input noise, and $\left\{t_0=0, t_1, \ldots, t_T=T_\mathrm{max}\right\}$ is a subsequence of timesteps $[0, \ldots, T_\mathrm{max}]$. $f_{\theta, t_j}$ is an intermediate stochastic sampling step involving denoising followed by renoising~\cite{dmd}: 
\begin{equation}
f_{\theta, t_j}\left(\bx_{t_j}^i\right)=\Psi\left({\bx}_{0|t_j}^i,\epsilon^i_{j-1}, t_{j-1}\right),
\label{eq:mid_denoise}
\end{equation}
where the denoising step is defined as 
\begin{equation}
{\bx}^i_{0|{t_{j}}}=G_\theta\left(\Psi\left({\bx}^i_{0|t_{j+1}},{\epsilon}^i_{t_{j}},t_j\right);\bx^{<i},t_j\right),
\label{eq:denoising_step}
\end{equation}
which denoises noisy input latents $\bx_{t_j}^i$ conditioned on clean historical frame chunks $\bx^{<i}$. Here, we omit the text condition of the denoising model for brevity. The intermediate path noise $\epsilon^i_{t_{j-1}}\sim\cN\left(0, I\right)$ is then sampled to introduce stochasticity and renoise the sample to the next timestep through the forward diffusion process~\cite{ho2020denoising,liu2022flow,lipman2022flow} 
\begin{equation}
\Psi\left(\bx,\epsilon_{t_{j}},t_{j}\right)=\alpha_{t_{j}} \bx+\sigma_{t_{j}} \epsilon_{t_{j}},
\label{eq:forward_diff}
\end{equation}
where $\alpha_{t_{j}}$ and $\sigma_{t_{j}}$ follow a predefined noise schedule over timestep $t$~\cite{lipman2022flow}. Such a renoising process is crucial to maintain sample fidelity of AR-VDMs (explained in \cref{sec:traj_noise_refine}).

\newparagraph{Step Distillation.} To achieve real-time responsiveness in streaming applications, the denoising model $G_\theta$ is typically step-distilled from a multi-step teacher model, allowing valid sampling with a small number of diffusion steps $T$. We mainly consider the state-of-the-art AR-VDM paradigm---Self-Forcing (SF)~\cite{selfforcing}. Unlike Teacher Forcing~\cite{teacherforcing} or Diffusion Forcing~\cite{diffusionforcing}, which condition on ground-truth history frames, SF conditions on self-rollout frames during training, thereby eliminating the train–test distribution gap. The history conditioning is achieved through causal attention and KV cache~\cite{causvid,selfforcing}. The self-rollout strategy further enables holistic optimization by directly evaluating the fidelity of generated clean videos, rather than relying on local score matching or flow matching. This is accomplished via the DMD loss~\cite{dmd} 
\begin{equation}
\nabla_\theta \mathcal{L}_{\mathrm{DMD}} = \mathbb{E}_t\left(\nabla_\theta \operatorname{KL}\left(q_{\theta , t}\left(\bx^{1: N}\right) \| p_{t}\left(\bx^{1: N}\right)\right)\right),
\label{eq:dmd}
\end{equation}
which approximates the gradient of the inverse KL divergence between the model and target data distributions.
\subsection{Learning Framework of AutoRefiner}
\label{sec:autorefiner}
Given a pre-trained AR-VDM $G_\theta$ and a differentiable scorer $R$ that measures video fidelity, our objective is to improve the sample quality of $G_\theta$ by steering its base distribution $q_\theta$ towards a tilted distribution 
\begin{equation}
q^\star_\theta\left(\bx^{1: N}\right)\propto q_\theta\left(\bx^{1: N}\right)\exp \left(R\left(\bx^{1: N}\right)\right),
\end{equation}
that better aligns $R$. In practice, $R$ can be instantiated in various forms, such as a distributional scorer like DMD loss in \cref{eq:dmd} or a human preference scorer~\cite{hpsv2,imagereward,internvid}.
To achieve this, rather than directly updating the base model parameters, we leverage the stochasticity~\cite{song2020score,anderson1982reverse} inherent in the stochastic sampling steps of \cref{eq:conditional}. Given the set of sampled noises, each rollout step from the conditional in \cref{eq:conditional} can be interpreted as a deterministic mapping from the sampled initial and intermediate noises to the output video chunk as follows
\begin{equation}
\bx^i=F_{\theta}\left(\bx_{t_T}^i,\mathbf{\epsilon}^i;\bx^{<i}\right)\sim q_\theta\left(\bx^i \mid \bx^{<i}\right),
\label{eq:determinstic_mapping}
\end{equation}
where $F_{\theta}$ denotes the composition of all denoising steps parameterized by $G_\theta$. Recent works show that the sampled noises in the diffusion process are not created equal~\cite{qi2024not,bon}. Given the same diffusion model, generation quality can be significantly improved by carefully selecting the noise samples. Inspired by these findings, we aim to learn the tilted distribution $q^\star_\theta\left(x^{1: N}\right)$ by transforming randomly sampled noises along the intermediate denoising path 
\begin{equation}
\bepsilon^{i\star}=\underset{\bepsilon^i}{\arg \max }\left(R\left(F_{\theta}\left(\bx_{t_T}^i,\bepsilon^i;\bx^{<i}\right)\right)-\operatorname{Reg}\left(\bepsilon^i\right)\right),
\label{eq:noise_opt}
\end{equation}
where $\bepsilon^i\sim\mathcal{N}\left(0, I\right)$, $\operatorname{Reg}\left(\bepsilon^i\right)$ is a regularization term that keeps the transformed noise close to the original one. \cref{eq:noise_opt} can be solved by per-sample noise optimization~\cite{reno} or searching~\cite{bon}. However, with a noise candidate number or optimization step of $L$, such methods have a time complexity of $O(TLN)$. Given that $L$ is usually hundreds or even thousands of steps, this is impractical for AR-VDMs, especially for real-time applications. Instead, we adopt the idea of  \textit{noise refiners}~\cite{ahn2024noise,goldennoise,hypernoise}. Specifically, we parameterize a feedforward network $T_\phi$ that learns to map a randomly sampled noise $\epsilon^i_{t_{j}}$ to an optimized one $\hat{\epsilon}^i_{t_{j}}=\epsilon^i_{t_{j}}+\Delta\epsilon^i_{t_{j}}$ with the following objective 
\begin{equation}
\mathcal{L}(\phi)=\mathbb{E}_{i,\epsilon^i,j}\left[R\left({\bx}^i_{0|{t_{j}}}\right) - \frac{1}{2}\left\|\Delta \epsilon^i_{t_{j}} \right\|^2\right],
\label{eq:noise_refine}
\end{equation}
where the random noise $\epsilon^i_{t_{j}}$ is refined by a predicted residual term from $T_\phi$ that conditions on both the clean history frames $\bx^{<i}$ and the denoising output from the previous timestep ${\bx}^i_{0|{t_j}}$, \ie, $\Delta \epsilon^i_{t_{j}}=T_\phi\left(\epsilon^i_{t_{j}};\bx^i_{0|t_{j+1}},\bx^{<i}\right)$. Following the denoising step in \cref{eq:denoising_step}, the interpolation between the refined noise $\hat{\epsilon}^i_{t_{j}}$ and the denoising output from the previous timestep $\bx^i_{0|t_{j+1}}$ is used as input to obtain the current-timestep denoising prediction via
\begin{equation}
{\bx}^i_{0|{t_{j}}}=G_\theta\left(\Psi\left({\bx}^i_{0|t_{j+1}},\hat{\epsilon}^i_{t_{j}},t_j\right);\bx^{<i},t_j\right),  
\end{equation}
for which the fidelity score $R\left({\bx}^i_{0|{t_{j}}}\right)$ is computed.
The $L_2$ term approximates the $\operatorname{KL}$ divergence between the modified noise distribution and the original Gaussian distribution~\cite{hypernoise}, which regularizes the refined noise to remain Gaussian distributed and ensures that the tilted distribution can maintain the sample diversity of the base model. 

\subsection{From Initial to Pathwise Noise Refinement}
\label{sec:traj_noise_refine}
The learning framework in \cref{sec:autorefiner} fundamentally differs from prior T2I noise refiners~\cite{ahn2024noise,goldennoise,hypernoise} with one crucial distinction: it refines the noise sequence $\left\{\epsilon_{1: T-1}^i\right\}$ sampled from the intermediate path, rather than the initial noise $\bx_{t_T}^i$. We next elaborate on the rationale behind this design.

\newparagraph{Limitation of Initial Noise Refinement.}
We first investigate the effect of applying the refiner to the initial noise space of AR-VDMs, following the design of prior text-to-image noise refiners~\cite{ahn2024noise,goldennoise,hypernoise}. Empirically, this strategy yields limited or unstable performance for AR-VDMs. Depending on the choice of the differentiable fidelity scorer \(R\), we observe two distinct failure behaviors:
\begin{itemize}
    \item \textbf{With distribution matching loss.} When \(R\) is instantiated as the DMD loss~\cite{dmd}, refining the initial noise $\bx_{t_T}^i$ yields moderate or worse performance (\cref{tab:dmd_comparison}). This finding contrasts with prior work that reports substantial improvements from optimization in the initial noise space for T2I diffusion models. We conjecture that this discrepancy arises from differences in sampling strategies. Earlier approaches typically adopt ODE-based samplers, where the initial noise constitutes the sole source of stochasticity; thus, optimizing it can yield inference-scaling behavior~\cite{bon}. In contrast, step-distilled AR-VDMs depend on stochastic sampling during few-step inference~\cite{dmd,consistencymodels}, as expressed in \cref{eq:mid_denoise}, where randomness is reintroduced at intermediate steps. The injection of such intermediate noise introduces additional randomness, which dilutes the influence of initial noise optimization. Consequently, optimizing the initial noise alone cannot reliably steer overall sample quality, as later stochastic perturbations partially overwrite the effects of the refined initialization.
    \item \textbf{With preference-based reward scores.}  
    In contrast, when \(R\) is instantiated as a preference-based reward scorer---aimed at improving perceptual quality and semantic alignment, we observe reward hacking behaviors such as motion degradation and grid-like artifacts (see \cref{fig:reward_comparison} and supplementary). We conjecture that this arises because the initial noise determines the low-frequency components of the generated video, which the refiner can easily exploit to obtain higher reward scores by taking shortcuts, \eg, the modified initial noise $\hat{\bx}_{t_T}^i$ can strongly influence the overall motion magnitude of the sample~\cite{howiwarpyournoise,burgert2025go}, enabling the refiner to trivially reconstructing the previous chunk, \ie, $F_{\theta}\left(\hat{\bx}_{t_T}^i,\mathbf{\epsilon}^i;\bx^{<i}\right)\approx \bx^{i-1}$. This shortcut increases the reward by producing static frames with minimal motion blur (see low motion degree in \cref{tab:reward_comparison} and examples in supplementary), rather than genuinely improving the quality of generated videos.
    \end{itemize}

\begin{figure}[t]    \centering\includegraphics[width=\linewidth]{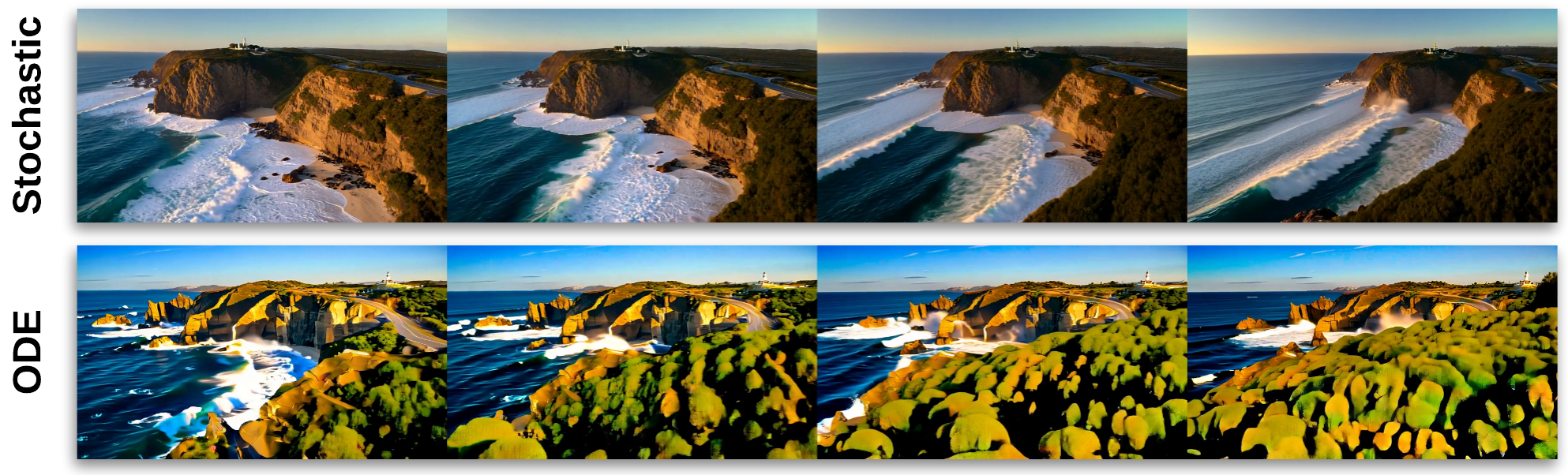}
\vspace{-6mm}
    \caption{Replacing the stochastic sampler of Self-Forcing with an ODE sampler leads to over-saturation and drifting artifacts.}
    \label{fig:ode_vs_stoch}
\end{figure}

\newparagraph{Pathwise Noise Refinement.} 
Given the limitations of initial noise refinement, we instead train $T_\phi$ with a different objective than previous methods: \textit{refinement along the stochastic sampling path}. Unlike prior T2I noise refiners that optimize the initial noise $\bx_{t_T}^i$, our formulation in \cref{eq:noise_refine} changes the refinement target to the intermediate noise sequence ${\epsilon^i_{t_{1:T-1}}}$. As shown in \cref{fig:ode_vs_stoch}, removing the stochasticity introduced by intermediate noises---by replacing the stochastic sampler with an ODE sampler---results in color saturation and drifting artifacts, underscoring its crucial role in preserving the fidelity of generated samples in AR-VDMs. Since intermediate noises perturb the denoised output from the previous timestep via the forward diffusion process~\cref{eq:forward_diff}, we conjecture that they act as a self-correction mechanism~\cite{karras2022elucidating}, injecting stochasticity to correct errors accumulated from earlier steps. Our learning framework in \cref{sec:autorefiner} leverages this property by training $T_\phi$ to steer the denoising path towards higher fidelity modes. 
\begin{figure}[t]    \centering\includegraphics[width=\linewidth]{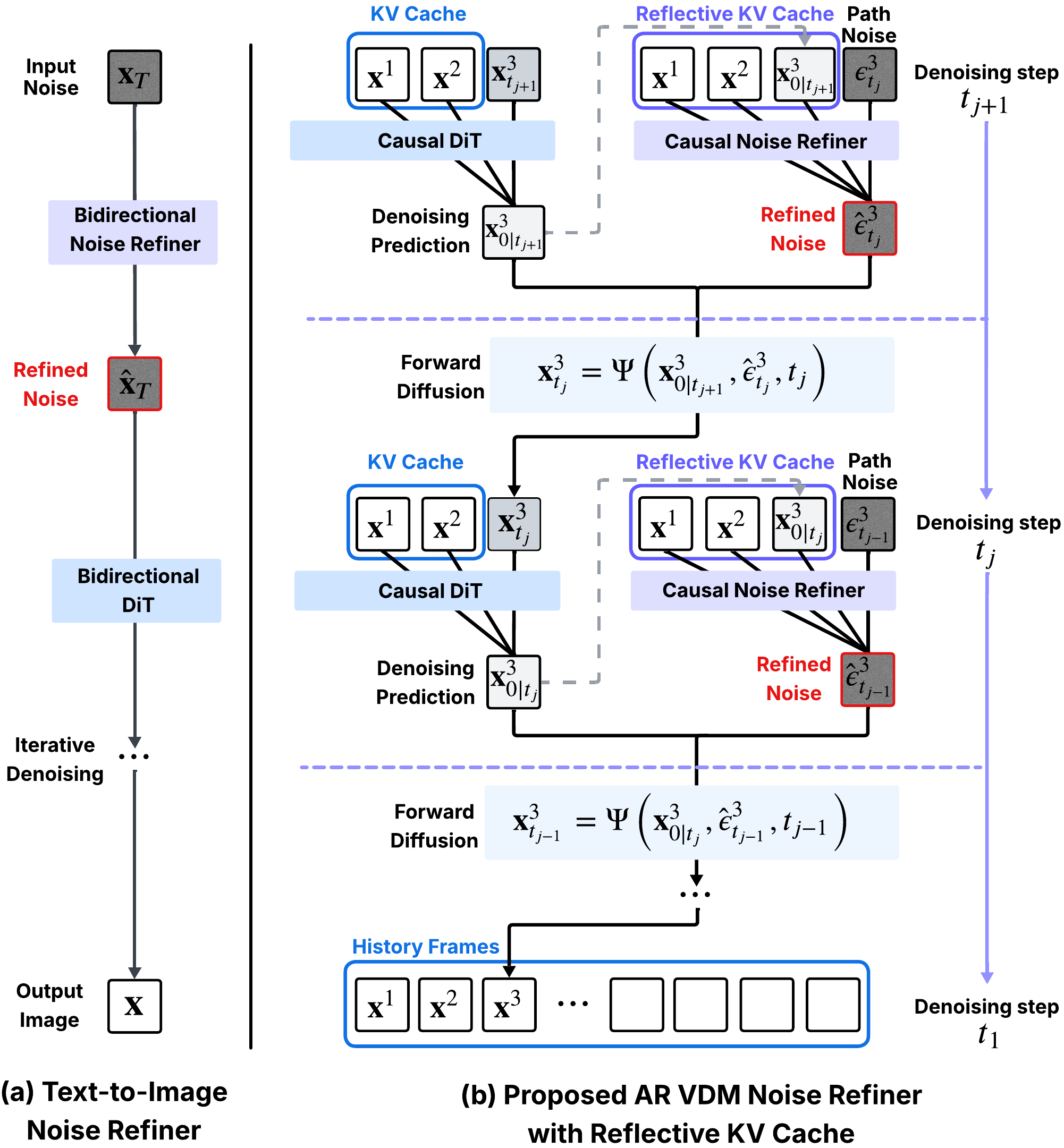}
    \caption{Illustration of the difference between existing T2I noise refiners and \name. In contrast to prior text-to-image (T2I) noise refiners that operate on the initial noise space, \name performs refinement along the intermediate denoising trajectory. Equipped with a \textbf{reflective KV cache}, \name conditions each noise refinement step on both historical frames and the denoised latent from the previous diffusion timestep.}
    \label{fig:ref_kv}
\end{figure}
\begin{algorithm}[t]
\caption{Training algorithm of \name}
\label{alg:finetune_inference}
\begin{algorithmic}[1]
\Require Denoising timesteps $\left\{t_1, \ldots, t_{T}\right\}$. Number of training video chunks $N$. Pre-trained AR-VDM $G_\theta$. Differentiable fidelity scorer $R$. 
\State Initialize noise refiner $T_\phi(\cdot)=\mathbf{0}$ via LoRA on $G_\theta$ 
\While{training} 
\State Initialize output $\bX \leftarrow[]$, $\mathcal{L}_{\text{reg}}\leftarrow0$.
\State Sample $s \sim \operatorname{Uniform}(1,2, \ldots, T-1)$
\For{$i = 1,...,N$} 
    \State Disable gradient computation
    \State Initialize $\bx_{t_T}^i \sim \cN(0,\bI)$
    \State ${\bx}_{0|{t_T}}^i \leftarrow G_\theta\left(\bx_{t_T}^i ; t_T, \bX\right)$
    \For{$j = T-1,...,s$}
            \State Enable gradient computation if $j=s$ 
            \State Sample $\epsilon^i_{t_{j}}\sim \cN(0,\bI)$
            \State $\Delta \epsilon^i_{t_{j}}\leftarrow T_\phi\left(\epsilon^i_{t_{j}};\bx^i_{0|t_{j+1}},\bX\right)$
             \State $\bx_{t_j}^i\leftarrow\Psi\left({\bx}^i_{0|t_{j+1}},{\epsilon}^i_{t_{j}}+\Delta \epsilon^i_{t_{j}},t_j\right)$
            \State ${\bx}_{0|{t_j}}^i \leftarrow G_\theta\left(\bx_{t_j}^i ; t_j, \bX\right)$

    \EndFor
    \State $\bX.\text{append}\left({\bx}_{0|{t_s}}^i\right)$
    \State $\mathcal{L}_{\text{reg}}\leftarrow \mathcal{L}_{\text{reg}}+\frac{1}{2}\left\|\Delta \epsilon^i_{t_{j}} \right\|^2$
\EndFor
\State Gradient step on $-\nabla_\phi\left({R\left(\bX\right)-
\mathcal{L}_\text{reg}}\right)$
\EndWhile
\end{algorithmic}
\end{algorithm}
\subsection{AR Noise Refiner with Reflective KV Cache}
\label{sec:ref_kv}

\newparagraph{Noise Refiner Network.} 
We initialize $T_\phi$ by the pre-trained autoregressive VDM $G_\theta$ to leverage its powerful prior knowledge. We train the refiner via LoRA~\cite{hu2021lora} adapters, which reduces memory overhead and avoids creating a separate copy for the base model. Following \citet{hypernoise}, we modify both the final layer weight and bias parameters of the base model to zero, which ensures that the initial output residual of $T_\phi$ is zero. This stabilizes training during the early iterations and prevents divergence.

\newparagraph{Reflective KV Cache.}
One fundamental advantage of autoregressive models is their ability to memorize past states via KV cache, allowing the models to predict future states based on their own previous predictions. We extend this capability to the autoregressive noise refiner $T_\phi$ with a reflective KV cache mechanism, which enables each prediction step of $T_\phi$ to condition on a richer context beyond the text-only conditioning used in prior T2I noise refiners. This includes two primary components. First, it inherits the conditioning on predicted history frames $\bx^{<i}$ from the base AR-VDM, ensuring that noise modulation does not destroy the overall temporal consistency. At the same time, this conditioning enables $T_\phi$ to enhance motion quality when optimized with a temporal fidelity scorer. Moreover, it also conditions on the denoising output from the previous timestep $\bx^i_{0|t_{j+1}}$, which also serves as the latent to be perturbed by the modulated noise through $\Psi\left({\bx}^i_{0|t_{j+1}},\hat{\epsilon}^i_{t_{j}},t_j\right)$. By conditioning on $\bx^i_{0|t_{j+1}}$, $T_\phi$ can refine intermediate path noises to correct errors or undesired artifacts present in $\bx^i_{0|t_{j+1}}$ that have accumulated from previous denoising steps. This process functions analogously to self-reflection~\cite{selfreflection}, wherein the model ``looks back" at its prior decisions to identify and rectify earlier mistakes. Conditioning on the reflective KV cache, $T_\phi$ takes in a randomly sampled intermediate path noise $\epsilon^i_{t_{j}}$ as input, with which the regularized optimization objective in \cref{eq:noise_refine} encourages the refiner to identify a locally optimal noise around $\epsilon^i_{t_{j}}$, avoiding distribution shift and preserving the stochasticity of the base model $G_\theta$, which is conceptually similar to the first-order noise search~\cite{bon}. To establish a consistent correspondence between each refined noise frame and its associated denoised latent, we apply the same Rotary Position Encoding (RoPE)~\cite{rope} to both $\epsilon^i_{t_{j}}$ and $\bx^i_{0|t_{j+1}}$.

\newparagraph{Efficient Optimization.}
Although $G_\theta$ is step-distilled, few-step sampling still incurs significant memory overhead due to maintaining the computational graph of the denoising chain $f_{\theta, t_1} \circ f_{\theta, t_2} \circ \ldots \circ f_{\theta, t_{T-1}}$. To address this, as shown in \cref{alg:finetune_inference}, we adopt a similar gradient truncation strategy from \citet{selfforcing}, which restricts backpropagation to a single, randomly selected final noise refinement step $t_s$, where $s \sim \operatorname{Uniform}(1,2, \ldots, T-1)$. We additionally detach the gradient flow from the predicted history frames and previous denoising output through the reflective KV cache.

\section{Experiments}
\label{sec:experiments}
In this section, we validate the performance of \name and the effect of the proposed components through quantitative and qualitative evaluations with three perspectives: 
\begin{itemize}
    \item \textit{Q1.} Does \name enhance the general generative capability of base AR-VDMs when trained with a generative distribution matching loss? 
    \item \textit{Q2.} Can \name improve specific preference dimensions when trained with reward-based objectives? 
    \item \textit{Q3.} Demonstrate the effectiveness of the proposed pathwise noise refinement and reflective KV cache.
\end{itemize}

\newparagraph{Datasets.} The training of the proposed method does not require any video data. We adopt the VidProS subset from VidProM~\cite{vidprom} as the training prompt dataset, which is filtered and preprocessed by \citet{selfforcing}. For evaluation, we use 946 prompts from VBench~\cite{huang2024vbench}. Please refer to supplementary for further implementation details.
\begin{figure*}[!ht]
    \centering
        \includegraphics[width=\linewidth]{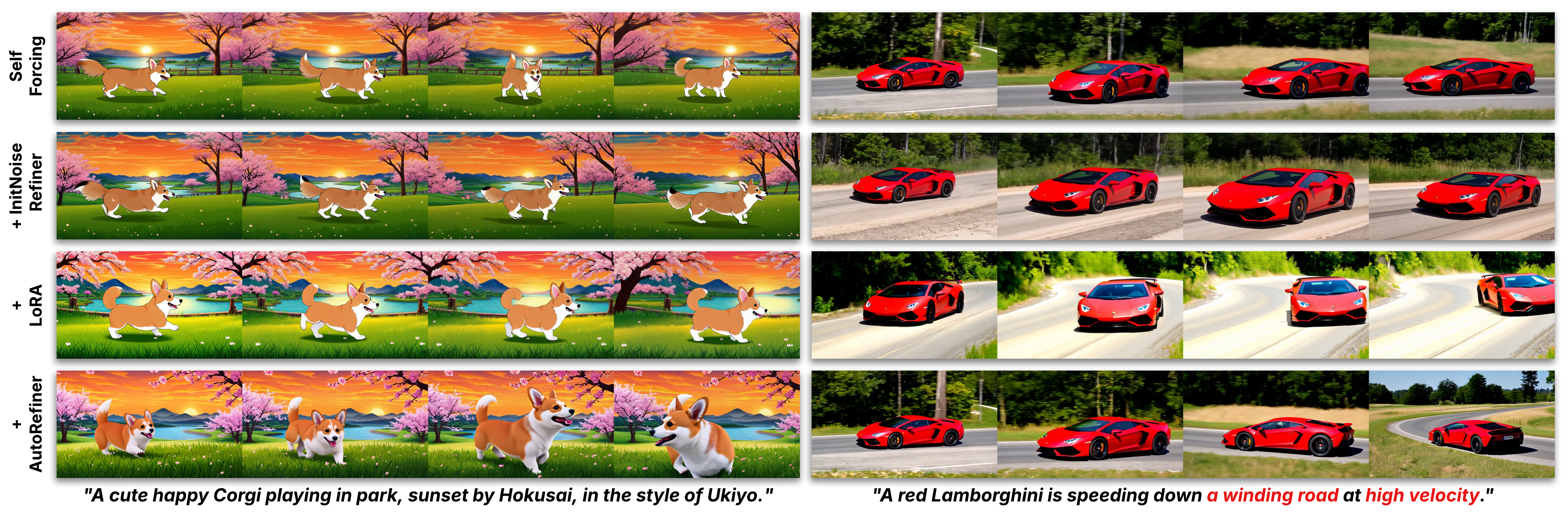}
        \vspace{-6mm}
        \caption{Qualitative comparison with baselines on training with DMD loss, using Self-Forcing~\cite{selfforcing} as the base model. Compared to the baselines, \name shows improved visual fidelity (as illustrated by the higher visual quality of the \textit{'corgi'} example), and motion alignment with text prompt (as shown by the faithfully reflected winding road and high velocity in the \textit{'car'} example).} 
        \label{fig:dmd_sf_comparison}
\end{figure*}
\begin{figure*}[ht]
    \centering
        \includegraphics[width=\linewidth]{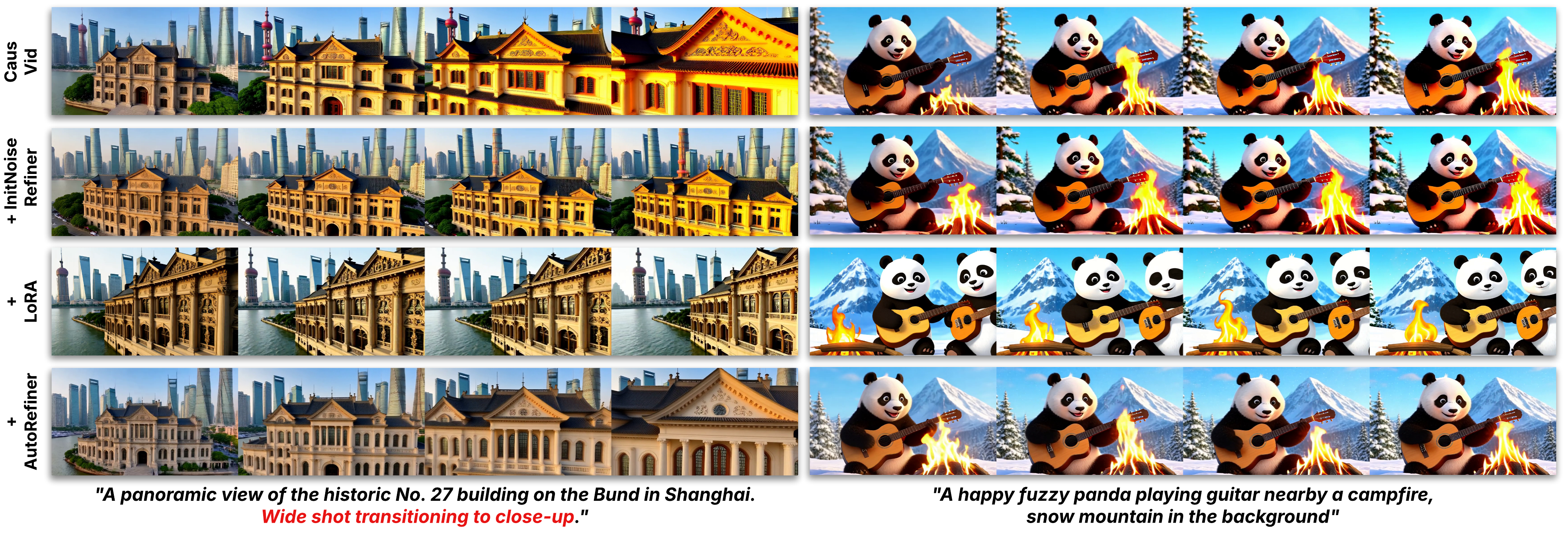}
        \vspace{-6mm}
        \caption{Qualitative comparison with baselines on training with DMD loss, using CausVid~\cite{causvid} as the base model. \name alleviates the cross-frame inconsistency issue of CausVid, as observed in the gradually saturated color of the ``\textit{building}" (left) and the ``\textit{sky}" (right).}
        \label{fig:dmd_cv_comparison} 
\end{figure*}
\subsection{Experiments with DMD-based Training}
\label{sec:exp_dmd}
For our experimental setup to answer \textit{Q1}, we train \name using the Distribution Matching Distillation (DMD) loss~\cite{dmd} as defined in \cref{eq:dmd}, and assess its effect on the general generative capability of the base AR-VDMs. Since the base AR-VDMs are already optimized with the DMD loss, we train \name as a plug-in module to further reduce the gap between generated and real data distributions. 

\newparagraph{Evaluation Metric.} Since this setting targets improving general video generation capability, we report overall video fidelity scores averaged across all VBench~\cite{huang2024vbench} dimensions.

\newparagraph{Baselines.} We consider the following baselines: 

\begin{table}[t]
\centering
\resizebox{\linewidth}{!}{
\footnotesize
\begin{tabular}{l c c c c}
    \toprule
    & \makecell[c]{{Num}\\ {Params} $\downarrow$} & \makecell[c]{{Total}\\ {Score} $\uparrow$} & \makecell[c]{{Quality}\\ {Score} $\uparrow$}
    & \makecell[c]{Semantic\\Score. $\uparrow$} \\
    \midrule
    \rowcolor{lightgray}Self-Forcing~\cite{selfforcing} & 1.3B & 84.05 & 84.93 & 80.54 \\
    +LoRA & 1.4B & 82.10 & 82.65 & 79.93 \\
    +InitNoiseRefiner & 1.4B & 83.76 & 84.63 & 80.29 \\
    \rowcolor{lightblue}+AutoRefiner (ours)  & 1.4B & \textbf{84.72} & \textbf{85.41} & \textbf{81.95} \\
    \midrule 
    \rowcolor{lightgray}CausVid~\cite{causvid}  & 1.3B & 83.06 & 84.24 & 78.35 \\
    +LoRA  & 1.4B & 83.43 & 84.74 & 78.18 \\
    +InitNoiseRefiner  & 1.4B & 82.23 & 83.13 & 78.64 \\
    \rowcolor{lightblue}+AutoRefiner (ours)  & 1.4B & \textbf{83.84} & \textbf{84.82} & \textbf{79.93}\\
    \bottomrule
\end{tabular}}
\caption{Comparison with baseline methods using distribution-matching distillation (DMD~\cref{eq:dmd}) loss as the training objective.
We compare our approach against baseline methods using overall video fidelity scores from VBench (best result in \textbf{bold}).}
\vspace{-4mm}

\label{tab:dmd_comparison}
\end{table}
\begin{itemize}
    \item \textbf{Baseline AR-VDMs.} We consider two most representative AR-VDMs as baselines---\textbf{Self-Forcing}~\cite{selfforcing} and \textbf{CausVid}~\cite{causvid} (based on Diffusion-Forcing~\cite{diffusionforcing}). \name is served as a plug-in noise refiner to both models.
    \item \textbf{Fine-tuning methods.} We also compare with fine-tuning via \textbf{LoRA} 
    and an initial-timestep noise refiner (\textbf{InitNoiseRefiner}) variant that trains a noise refiner for only the initial noise $\bx_{t_T}^i$, which follows the similar implementation of \citet{hypernoise} for T2I models. 
\end{itemize}
\newparagraph{Results.} \cref{tab:dmd_comparison} shows the quantitative results on overall VBench scores. For Self-Forcing, LoRA degrades performance, indicating that simply adding more parameters through fine-tuning does not improve generation fidelity. Likewise, InitNoiseRefiner fails to yield the performance gains observed in prior T2I works. In contrast, \name improves both overall quality and semantic scores, as also evidenced qualitatively in \cref{fig:dmd_sf_comparison}. 
For CausVid, LoRA degrades textual alignment---\eg, incorrect camera motion in \cref{fig:dmd_sf_comparison} (left) and an incorrect number of objects in \cref{fig:dmd_sf_comparison} (right), while the initial noise refiner lowers visual quality (\cref{tab:dmd_comparison}). In contrast, \name achieves clear gains for both dimensions. As shown in \cref{fig:dmd_cv_comparison}, CausVid tends to exhibit over-saturation as more frames are generated. We conjecture that the cause is due to compounding errors accumulated during rollout. \name reduces such artifacts and improves the overall cross-frame consistency.
\subsection{Experiments with Reward-based Training}
\label{sec:exp_reward}
To evaluate \textit{Q2}, we train \name using a combination of reward models following \citet{li2024t2v}: two image reward models---HPSv2.1~\cite{hpsv2}, which captures human preference, and CLIPScore~\cite{CLIP}, which measures frame-level text alignment, and a video reward model---InternVideo2~\cite{internvideo2}, which measures spatiotemporal consistency with text prompts. Under this setting, \name is trained to steer the pre-trained AR-VDMs toward preference dimensions emphasizing perceptual quality and semantic alignment.

\newparagraph{Evaluation Metric.} In accordance with the dimensions emphasized by reward models, we evaluate performance on eight metrics from VBench focusing on (1) \textit{perceptual quality} (imaging and aesthetic quality; motion degree and style) and (2) \textit{semantic alignment} (textual alignment measured by Tag2Text~\cite{tag2text} and ViClip~\cite{internvid}; object class accuracy for both single- and multi-object generation). We also report the introduced inference-time computational overhead relative to the base model, including additional NFE ($\Delta\text{NFE}$), reward verification (\#Verify), VAE decoding ($\Delta\text{VAE}$), and the resulting frames-per-second (FPS). 
\begin{figure*}[ht]
    \centering
        \includegraphics[width=\linewidth]{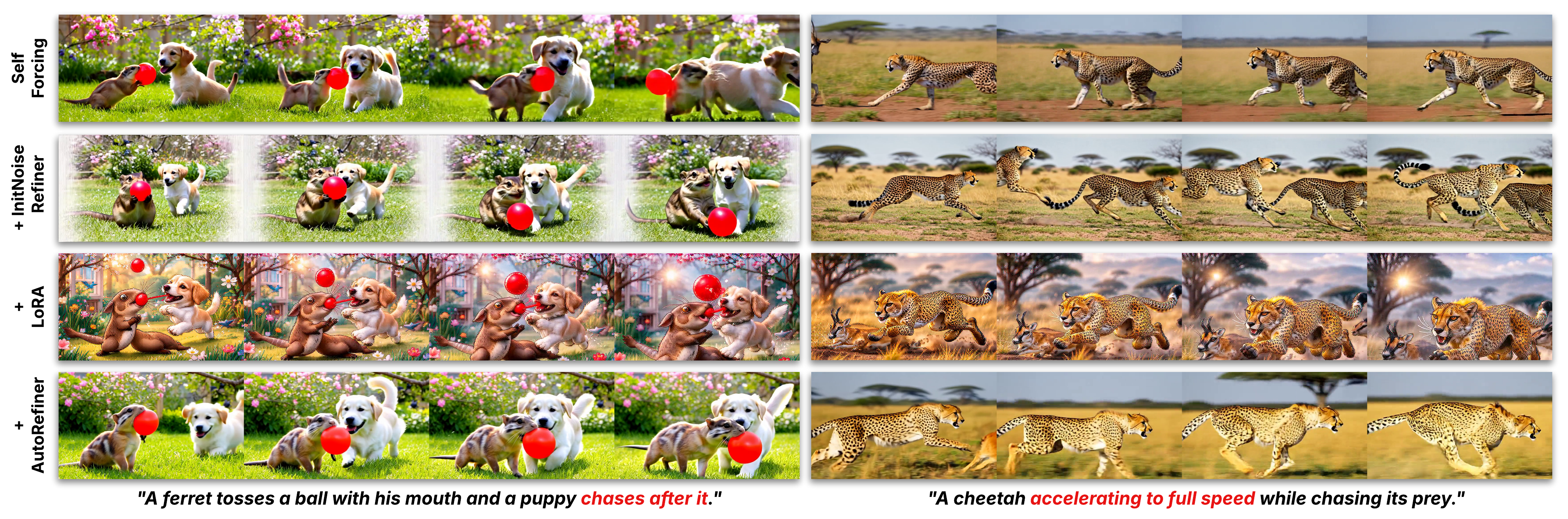}
        \vspace{-2mm}
        \caption{Qualitative comparison with baselines on training with reward scores. \name improves perceptual quality (left example) and temporal coherence (right example), without suffering from reward hacking issues observed in LoRA and initial noise refiner.}
        \label{fig:reward_comparison}
\end{figure*}
\begin{table*}[t]
\centering
    \resizebox{\linewidth}{!}{\centering
    \footnotesize
    \begin{tabular}{l c c c c | c c c c c c c c c c c }
        \toprule  
        \multirow{2}{*}{Method} & \multicolumn{4}{c}{Compute Overhead} & \multicolumn{2}{c}{Visual Quality$\uparrow$} & \multicolumn{2}{c}{Motion Quality$\uparrow$} & \multicolumn{2}{c}{Textual Alignment$\uparrow$} & \multicolumn{2}{c}{Object Accuracy$\uparrow$}\\
        \cmidrule(lr){2-5} \cmidrule(lr){6-7} \cmidrule(lr){8-9} \cmidrule(lr){10-11} \cmidrule(lr){12-13}
        & $\Delta\text{NFE}\downarrow$ & \#Verify$\downarrow$  & $\Delta\text{VAE}\downarrow$ & FPS$\uparrow$ & Imaging & Aesthetic & Degree & Style & Tag2Text & ViClip & Single & Multi \\
        \midrule
        \rowcolor{lightgray}Self-Forcing~\cite{selfforcing}  & - & - & - & 10.3 & 69.24 & 65.78 & \secondbest{69.44} & 24.58 & 52.83 & 26.80 & 94.22 & 88.41 \\
        + BoN  & 112 & 5 & 4 & 2.48 & 68.76 & 66.03 & 63.89 & 24.59 & \best{60.68} & \secondbest{27.13} & 92.41 & \best{91.84} \\
        + SoP & 112 & 20 & 19 & 1.22 & 68.77 & 66.03 & 63.89 & \secondbest{24.60} & \secondbest{60.47} & \best{27.14} & 92.48 & \secondbest{91.62} \\
        + LoRA  & 0 & 0 & 0 & 9.48 & \best{74.52} & \best{67.30} & 0.00 & 22.76 & 50.73 & 26.12 & \secondbest{94.78} & 82.62 \\
        + InitNoiseRefiner  & 7 & 0 & 0 & 7.57 & 70.76 & 66.04 & 30.56 & 24.40 & 54.00 & 26.70 & \best{94.94} & 83.08 \\ 
        \rowcolor{lightblue}+ AutoRefiner (ours)  & 21 & 0 & 0 & 6.32 & \secondbest{70.89} & \secondbest{66.26} & \best{79.17} & \best{24.68} & {57.63} & {26.94} & {94.62} & {91.23} \\ 
        \bottomrule
    \end{tabular}}
    \caption{Comparison with baseline methods using reward models as objective/guidance, the \best{best} and \textcolor{mediumslateblue}{\textbf{second best}} results are highlighted.}
    \label{tab:reward_comparison}
\vspace{-5mm}
\end{table*}
\newparagraph{Baselines.} We use Self-Forcing as the base model under this setting. In addition to LoRA and InitNoiseRefiner introduced in \cref{sec:exp_dmd}, we further compare against two searching-based inference-time alignment methods adapted from text-to-image works such as \citet{bon}:
\begin{itemize}
    \item \textbf{Best-of-N (BoN).} Randomly samples 5 initial noise candidates for each video chunk and selects the one achieving the highest reward score as output. 
    \item \textbf{Search-over-Path (SoP).} Samples five noise candidates at each initial and intermediate timestep, choosing after every step the denoised output with the highest reward to serve as the input to the next timestep.
\end{itemize}
\newparagraph{Results.} The quantitative results of reward-based training are shown in \cref{tab:reward_comparison}. \name consistently improves Self-Forcing across all evaluation dimensions, indicating effective alignment with the reward models---also reflected in the improved perceptual quality and motion coherence in \cref{fig:reward_comparison}. Both LoRA and InitNoiseRefiner exhibit clear reward hacking behaviors with severely degraded motion degree and visual artifacts such as grid patterns (InitNoiseRefiner) or cartoonish textures (LoRA). Although BoN and SoP achieve noticeable gains in semantic alignment, they incur substantial computational overhead, whereas \name requires considerably less inference computation. 
\subsection{Ablation studies}
\newparagraph{Ablation on Refined Timesteps.} 
We conduct an ablation over different intermediate refinement timesteps $\{750, 500, 250\}$ to analyze the contribution of each timestep. We consider two settings: (1) refining a single timestep, and (2) omitting one of the three intermediate timesteps. The results are shown in \cref{tab:ablation_timestep}. We observe: (1) earlier timesteps contribute more to performance gains. We conjecture that this is because the denoising prediction ${\bx}^i_{0|{t_{j}}}$ at earlier timesteps has lower sample fidelity, which leaves the noise refiner more room for correction; (2) omitting any step leads to a degradation in performance. Based on these findings, we apply noise refinement for all three intermediate denoising steps.
\begin{table}[ht]
\centering
\resizebox{\linewidth}{!}{
\footnotesize
\begin{tabular}{ccccccc}
    \toprule  
    \multicolumn{3}{c}{Refined Timesteps} & \multirow{2}{*}{Quality Score $\uparrow$} & \multirow{2}{*}{Semantic Score $\uparrow$} & \multirow{2}{*}{Total Score $\uparrow$} \\
    \cmidrule(lr){1-3}
    $750$ & $500$ & $250$ &  \\
    \midrule
    \cmark & \xmark & \xmark & 85.34 & 81.68 & 84.61 \\
    \xmark & \cmark & \xmark & 85.17 & 81.53 & 84.43 \\
    \xmark & \xmark & \cmark & 85.16 & 81.32 & 84.40 \\
    \midrule
    \xmark & \cmark & \cmark & 85.15 & 81.76 & 84.48 \\
    \cmark & \xmark & \cmark & 85.33 & 81.73 & 84.61 \\
    \cmark & \cmark & \xmark & 85.21 & 81.72 & 84.51 \\
    \cmark & \cmark & \cmark & \textbf{85.41} & \textbf{81.95} & \textbf{84.72} \\
    \bottomrule
\end{tabular}}
\caption{Ablation on diffusion timesteps used for refinement.}
\label{tab:ablation_timestep}
\end{table}
\begin{table}[ht]
\centering
\resizebox{\linewidth}{!}{
\footnotesize
\begin{tabular}{ccccccc}
    \toprule  
    \multicolumn{2}{c}{KV cache form} & \multirow{2}{*}{Quality Score $\uparrow$} & \multirow{2}{*}{Semantic Score $\uparrow$} & \multirow{2}{*}{Total Score $\uparrow$} \\
    \cmidrule(lr){1-2}
    History & Denoised \\
    \midrule
    \xmark & \xmark & 85.13 & 81.35 & 84.37 \\
    \xmark & \cmark & 85.11 & 81.57 & 84.41 \\
    \cmark & \xmark & 85.17 & 81.24 & 84.39 \\
    \cmark & \cmark & \textbf{85.41} & \textbf{81.95} & \textbf{84.72}\\
    \bottomrule
\end{tabular}}
\caption{Ablation on different KV cache mechanisms.}
\label{tab:ablation_kv}
\end{table}

\newparagraph{Ablation on KV Cache Mechanisms.}
We ablate the proposed reflective KV cache mechanism (\cref{sec:ref_kv}) by selectively removing its two conditioning components: the history frames $\bx^{<i}$, the denoised output from the previous timestep $\bx^i_{0|t_{j+1}}$. We find that incorporating both conditions yields the best performance, validating our design choice.

\section{Conclusion}
\label{sec:conclusion}
We introduced \name, an efficient noise refinement framework for AR-VDMs. Unlike prior T2I refiners that operate on initial noise, \name performs pathwise refinement along the denoising trajectory and integrates a reflective KV-cache to leverage temporal and contextual cues.
Across both distributional and reward-based objectives, \name consistently improves sample fidelity, motion quality, and semantic alignment of base AR-VDMs. We hope this work inspires future research on efficient inference-time alignment for improved video generation.

{   
    \small
    \bibliographystyle{ieeenat_fullname}
    \bibliography{main}

@String(NIPS= {Adv. Neural Inform. Process. Syst.})

@String(ICLR = {Int. Conf. Learn. Represent.})

@String(NIPS  = {NeurIPS})

@String(ICLR  = {ICLR})

@String(ICML  = {ICML})

@inproceedings{brown2020language,
  title={Language models are few-shot learners},
  author={Brown, Tom and Mann, Benjamin and Ryder, Nick and Subbiah, Melanie and Kaplan, Jared D and Dhariwal, Prafulla and Neelakantan, Arvind and Shyam, Pranav and Sastry, Girish and Askell, Amanda and others},
  booktitle=nips,
  volume={33},
  pages={1877--1901},
  year={2020}
}

@article{classifier_free_guidance,
  author       = {Jonathan Ho and
                  Tim Salimans},
  title        = {Classifier-Free Diffusion Guidance},
  journal      = {CoRR},
  volume       = {abs/2207.12598},
  year         = {2022},
  url          = {https://doi.org/10.48550/arXiv.2207.12598},
  doi          = {10.48550/arXiv.2207.12598},
  eprinttype    = {arXiv},
  eprint       = {2207.12598},
  bibsource    = {dblp computer science bibliography, https://dblp.org}
}

@inproceedings{hu2021lora,
  author       = {Edward J. Hu and
                  Yelong Shen and
                  Phillip Wallis and
                  Zeyuan Allen{-}Zhu and
                  Yuanzhi Li and
                  Shean Wang and
                  Lu Wang and
                  Weizhu Chen},
  title        = {LoRA: Low-Rank Adaptation of Large Language Models},
  booktitle    = {{ICLR}},
  publisher    = {OpenReview.net},
  year         = {2022}
}

@article{wang2023prolificdreamer,
  title={ProlificDreamer: High-Fidelity and Diverse Text-to-3D Generation with Variational Score Distillation},
  author={Wang, Zhengyi and Lu, Cheng and Wang, Yikai and Bao, Fan and Li, Chongxuan and Su, Hang and Zhu, Jun},
  journal={arXiv preprint arXiv:2305.16213},
  year={2023}
}

@inproceedings{song2020score,
  author       = {Yang Song and
                  Jascha Sohl{-}Dickstein and
                  Diederik P. Kingma and
                  Abhishek Kumar and
                  Stefano Ermon and
                  Ben Poole},
  title        = {Score-Based Generative Modeling through Stochastic Differential Equations},
  booktitle    = {{ICLR}},
  publisher    = {OpenReview.net},
  year         = {2021}
}

@inproceedings{ho2020denoising,
  title={Denoising diffusion probabilistic models},
  author={Ho, Jonathan and Jain, Ajay and Abbeel, Pieter},
  booktitle=nips,
  volume={33},
  pages={6840--6851},
  year={2020}
}

@inproceedings{CLIP,
  title={Learning transferable visual models from natural language supervision},
  author={Radford, Alec and Kim, Jong Wook and Hallacy, Chris and Ramesh, Aditya and Goh, Gabriel and Agarwal, Sandhini and Sastry, Girish and Askell, Amanda and Mishkin, Pamela and Clark, Jack and others},
  booktitle=icml,
  pages={8748--8763},
  year={2021},
  organization={PMLR}
}

@article{karras2022elucidating,
  title={Elucidating the design space of diffusion-based generative models},
  author={Karras, Tero and Aittala, Miika and Aila, Timo and Laine, Samuli},
  journal={Advances in Neural Information Processing Systems},
  volume={35},
  pages={26565--26577},
  year={2022}
}

@article{wang2023lavie,
  title={Lavie: High-quality video generation with cascaded latent diffusion models},
  author={Wang, Yaohui and Chen, Xinyuan and Ma, Xin and Zhou, Shangchen and Huang, Ziqi and Wang, Yi and Yang, Ceyuan and He, Yinan and Yu, Jiashuo and Yang, Peiqing and others},
  journal={arXiv preprint arXiv:2309.15103},
  year={2023}
}

@article{he2022latent,
  title={Latent video diffusion models for high-fidelity long video generation},
  author={He, Yingqing and Yang, Tianyu and Zhang, Yong and Shan, Ying and Chen, Qifeng},
  journal={arXiv preprint arXiv:2211.13221},
  year={2022}
}

@article{ho2022video,
  title={Video diffusion models},
  author={Ho, Jonathan and Salimans, Tim and Gritsenko, Alexey and Chan, William and Norouzi, Mohammad and Fleet, David J},
  journal={Advances in Neural Information Processing Systems},
  volume={35},
  pages={8633--8646},
  year={2022}
}

@misc{wang2023modelscopetexttovideotechnicalreport,
      title={ModelScope Text-to-Video Technical Report}, 
      author={Jiuniu Wang and Hangjie Yuan and Dayou Chen and Yingya Zhang and Xiang Wang and Shiwei Zhang},
      year={2023},
      eprint={2308.06571},
      archivePrefix={arXiv},
      primaryClass={cs.CV},
      url={https://arxiv.org/abs/2308.06571}, 
}

@article{ho2022imagen,
  title={Imagen video: High definition video generation with diffusion models},
  author={Ho, Jonathan and Chan, William and Saharia, Chitwan and Whang, Jay and Gao, Ruiqi and Gritsenko, Alexey and Kingma, Diederik P and Poole, Ben and Norouzi, Mohammad and Fleet, David J and others},
  journal={arXiv preprint arXiv:2210.02303},
  year={2022}
}

@article{singer2022make,
  title={Make-a-video: Text-to-video generation without text-video data},
  author={Singer, Uriel and Polyak, Adam and Hayes, Thomas and Yin, Xi and An, Jie and Zhang, Songyang and Hu, Qiyuan and Yang, Harry and Ashual, Oron and Gafni, Oran and others},
  journal={arXiv preprint arXiv:2209.14792},
  year={2022}
}

@article{zhang2023show,
  title={Show-1: Marrying pixel and latent diffusion models for text-to-video generation},
  author={Zhang, David Junhao and Wu, Jay Zhangjie and Liu, Jia-Wei and Zhao, Rui and Ran, Lingmin and Gu, Yuchao and Gao, Difei and Shou, Mike Zheng},
  journal={arXiv preprint arXiv:2309.15818},
  year={2023}
}

@inproceedings{chen2024videocrafter2,
  title={Videocrafter2: Overcoming data limitations for high-quality video diffusion models},
  author={Chen, Haoxin and Zhang, Yong and Cun, Xiaodong and Xia, Menghan and Wang, Xintao and Weng, Chao and Shan, Ying},
  booktitle={Proceedings of the IEEE/CVF Conference on Computer Vision and Pattern Recognition},
  pages={7310--7320},
  year={2024}
}

@article{guo2023animatediff,
  title={Animatediff: Animate your personalized text-to-image diffusion models without specific tuning},
  author={Guo, Yuwei and Yang, Ceyuan and Rao, Anyi and Liang, Zhengyang and Wang, Yaohui and Qiao, Yu and Agrawala, Maneesh and Lin, Dahua and Dai, Bo},
  journal={arXiv preprint arXiv:2307.04725},
  year={2023}
}

@misc{Open-Sora,
  title={Open-Sora},
  author={HPC-AI Tech},
  howpublished ={\url{https://github.com/hpcaitech/Open-Sora}},
  year={2024}
}

@article{liu2022flow,
  title={Flow straight and fast: Learning to generate and transfer data with rectified flow},
  author={Liu, Xingchao and Gong, Chengyue and Liu, Qiang},
  journal={arXiv preprint arXiv:2209.03003},
  year={2022}
}

@article{SSIM,
  title={Image quality assessment: from error visibility to structural similarity},
  author={Wang, Zhou and Bovik, Alan C and Sheikh, Hamid R and Simoncelli, Eero P},
  journal={IEEE transactions on image processing},
  volume={13},
  number={4},
  pages={600--612},
  year={2004},
  publisher={IEEE}
}

@inproceedings{PSNR,
  title={Image quality metrics: PSNR vs. SSIM},
  author={Hore, Alain and Ziou, Djemel},
  booktitle={2010 20th international conference on pattern recognition},
  pages={2366--2369},
  year={2010},
  organization={IEEE}
}

@inproceedings{lpips,
  title={The unreasonable effectiveness of deep features as a perceptual metric},
  author={Zhang, Richard and Isola, Phillip and Efros, Alexei A and Shechtman, Eli and Wang, Oliver},
  booktitle={Proceedings of the IEEE conference on computer vision and pattern recognition},
  pages={586--595},
  year={2018}
}

@inproceedings{stablediffusion,
  title={High-resolution image synthesis with latent diffusion models},
  author={Rombach, Robin and Blattmann, Andreas and Lorenz, Dominik and Esser, Patrick and Ommer, Bj{\"o}rn},
  booktitle={Proceedings of the IEEE/CVF conference on computer vision and pattern recognition},
  pages={10684--10695},
  year={2022}
}

@misc{flux,
  title = {Flux},
  author = {Black-Forest},
  howpublished = {https://blackforestlabs.ai/announcing-black-forest-labs/},
  year = 2024
}

@inproceedings{imagen,
  title={Photorealistic text-to-image diffusion models with deep language understanding},
  author={Saharia, Chitwan and Chan, William and Saxena, Saurabh and Li, Lala and Whang, Jay and Denton, Emily L and Ghasemipour, Kamyar and Gontijo Lopes, Raphael and Karagol Ayan, Burcu and Salimans, Tim and others},
  booktitle    = {NeurIPS},
  year         = {2022}
}

@inproceedings{glide,
  author       = {Alexander Quinn Nichol and
                  Prafulla Dhariwal and
                  Aditya Ramesh and
                  Pranav Shyam and
                  Pamela Mishkin and
                  Bob McGrew and
                  Ilya Sutskever and
                  Mark Chen},
  title        = {{GLIDE:} Towards Photorealistic Image Generation and Editing with
                  Text-Guided Diffusion Models},
  booktitle    = {{ICML}},
  series       = {Proceedings of Machine Learning Research},
  volume       = {162},
  pages        = {16784--16804},
  publisher    = {{PMLR}},
  year         = {2022}
}

@article{dalle,
  title={Hierarchical text-conditional image generation with clip latents},
  author={Ramesh, Aditya and Dhariwal, Prafulla and Nichol, Alex and Chu, Casey and Chen, Mark},
  journal={arXiv preprint arXiv:2204.06125},
  volume={1},
  number={2},
  pages={3},
  year={2022}
}

@misc{mochi,
  title = {mochi},
  author = {genmo},
  howpublished = {https://www.genmo.ai/blog},
  year = 2024
}

@misc{hunyuan,
  title = {hunyuan},
  author = {tencent},
  howpublished = {https://aivideo.hunyuan.tencent.com/},
  year = 2024
}

@inproceedings{sd3,
  title={Scaling rectified flow transformers for high-resolution image synthesis},
  author={Esser, Patrick and Kulal, Sumith and Blattmann, Andreas and Entezari, Rahim and M{\"u}ller, Jonas and Saini, Harry and Levi, Yam and Lorenz, Dominik and Sauer, Axel and Boesel, Frederic and others},
  booktitle={Forty-first International Conference on Machine Learning},
  year={2024}
}

@misc{vae,
  title={Auto-encoding variational bayes},
  author={Kingma, Diederik P and Welling, Max and others},
  year={2013},
  publisher={Banff, Canada}
}

@article{wan2025wan,
  title={Wan: Open and advanced large-scale video generative models},
  author={Wan, Team and Wang, Ang and Ai, Baole and Wen, Bin and Mao, Chaojie and Xie, Chen-Wei and Chen, Di and Yu, Feiwu and Zhao, Haiming and Yang, Jianxiao and others},
  journal={arXiv preprint arXiv:2503.20314},
  year={2025}
}

@inproceedings{huang2024vbench,
  title={Vbench: Comprehensive benchmark suite for video generative models},
  author={Huang, Ziqi and He, Yinan and Yu, Jiashuo and Zhang, Fan and Si, Chenyang and Jiang, Yuming and Zhang, Yuanhan and Wu, Tianxing and Jin, Qingyang and Chanpaisit, Nattapol and others},
  booktitle={Proceedings of the IEEE/CVF Conference on Computer Vision and Pattern Recognition},
  pages={21807--21818},
  year={2024}
}

@article{yang2024cogvideox,
  title={Cogvideox: Text-to-video diffusion models with an expert transformer},
  author={Yang, Zhuoyi and Teng, Jiayan and Zheng, Wendi and Ding, Ming and Huang, Shiyu and Xu, Jiazheng and Yang, Yuanming and Hong, Wenyi and Zhang, Xiaohan and Feng, Guanyu and others},
  journal={arXiv preprint arXiv:2408.06072},
  year={2024}
}

@article{hong2022cogvideo,
  title={Cogvideo: Large-scale pretraining for text-to-video generation via transformers},
  author={Hong, Wenyi and Ding, Ming and Zheng, Wendi and Liu, Xinghan and Tang, Jie},
  journal={arXiv preprint arXiv:2205.15868},
  year={2022}
}

@article{chen2023videocrafter1,
  title={Videocrafter1: Open diffusion models for high-quality video generation},
  author={Chen, Haoxin and Xia, Menghan and He, Yingqing and Zhang, Yong and Cun, Xiaodong and Yang, Shaoshu and Xing, Jinbo and Liu, Yaofang and Chen, Qifeng and Wang, Xintao and others},
  journal={arXiv preprint arXiv:2310.19512},
  year={2023}
}

@article{song2025history,
  title={History-guided video diffusion},
  author={Song, Kiwhan and Chen, Boyuan and Simchowitz, Max and Du, Yilun and Tedrake, Russ and Sitzmann, Vincent},
  journal={arXiv preprint arXiv:2502.06764},
  year={2025}
}

@article{diffusionforcing,
  title={Diffusion forcing: Next-token prediction meets full-sequence diffusion},
  author={Chen, Boyuan and Mart{\'\i} Mons{\'o}, Diego and Du, Yilun and Simchowitz, Max and Tedrake, Russ and Sitzmann, Vincent},
  journal={Advances in Neural Information Processing Systems},
  volume={37},
  pages={24081--24125},
  year={2024}
}

@article{selfforcing,
  title={Self Forcing: Bridging the Train-Test Gap in Autoregressive Video Diffusion},
  author={Huang, Xun and Li, Zhengqi and He, Guande and Zhou, Mingyuan and Shechtman, Eli},
  journal={arXiv preprint arXiv:2506.08009},
  year={2025}
}

@article{framepack,
  title={Packing input frame context in next-frame prediction models for video generation},
  author={Zhang, Lvmin and Agrawala, Maneesh},
  journal={arXiv preprint arXiv:2504.12626},
  year={2025}
}

@article{skyreels,
  title={Skyreels-v2: Infinite-length film generative model},
  author={Chen, Guibin and Lin, Dixuan and Yang, Jiangping and Lin, Chunze and Zhu, Junchen and Fan, Mingyuan and Zhang, Hao and Chen, Sheng and Chen, Zheng and Ma, Chengcheng and others},
  journal={arXiv preprint arXiv:2504.13074},
  year={2025}
}

@article{magi,
  title={MAGI-1: Autoregressive Video Generation at Scale},
  author={Teng, Hansi and Jia, Hongyu and Sun, Lei and Li, Lingzhi and Li, Maolin and Tang, Mingqiu and Han, Shuai and Zhang, Tianning and Zhang, WQ and Luo, Weifeng and others},
  journal={arXiv preprint arXiv:2505.13211},
  year={2025}
}

@article{fifo,
  title={Fifo-diffusion: Generating infinite videos from text without training},
  author={Kim, Jihwan and Kang, Junoh and Choi, Jinyoung and Han, Bohyung},
  journal={Advances in Neural Information Processing Systems},
  volume={37},
  pages={89834--89868},
  year={2024}
}

@inproceedings{Ardiffusion,
  title={Ar-diffusion: Asynchronous video generation with auto-regressive diffusion},
  author={Sun, Mingzhen and Wang, Weining and Li, Gen and Liu, Jiawei and Sun, Jiahui and Feng, Wanquan and Lao, Shanshan and Zhou, SiYu and He, Qian and Liu, Jing},
  booktitle={Proceedings of the Computer Vision and Pattern Recognition Conference},
  pages={7364--7373},
  year={2025}
}

@inproceedings{rollingdiffusion,
  title={Large-scale reinforcement learning for diffusion models},
  author={Zhang, Yinan and Tzeng, Eric and Du, Yilun and Kislyuk, Dmitry},
  booktitle={European Conference on Computer Vision},
  pages={1--17},
  year={2024},
  organization={Springer}
}

@inproceedings{pavdm,
  title={Progressive autoregressive video diffusion models},
  author={Xie, Desai and Xu, Zhan and Hong, Yicong and Tan, Hao and Liu, Difan and Liu, Feng and Kaufman, Arie and Zhou, Yang},
  booktitle={Proceedings of the Computer Vision and Pattern Recognition Conference},
  pages={6322--6332},
  year={2025}
}

@article{kodaira2025streamdit,
  title={Streamdit: Real-time streaming text-to-video generation},
  author={Kodaira, Akio and Hou, Tingbo and Hou, Ji and Tomizuka, Masayoshi and Zhao, Yue},
  journal={arXiv preprint arXiv:2507.03745},
  year={2025}
}

@inproceedings{causvid,
  title={From slow bidirectional to fast autoregressive video diffusion models},
  author={Yin, Tianwei and Zhang, Qiang and Zhang, Richard and Freeman, William T and Durand, Fredo and Shechtman, Eli and Huang, Xun},
  booktitle={Proceedings of the Computer Vision and Pattern Recognition Conference},
  pages={22963--22974},
  year={2025}
}

@inproceedings{dmd,
  title={One-step diffusion with distribution matching distillation},
  author={Yin, Tianwei and Gharbi, Micha{\"e}l and Zhang, Richard and Shechtman, Eli and Durand, Fredo and Freeman, William T and Park, Taesung},
  booktitle={Proceedings of the IEEE/CVF conference on computer vision and pattern recognition},
  pages={6613--6623},
  year={2024}
}

@article{schmidt2019generalization,
  title={Generalization in generation: A closer look at exposure bias},
  author={Schmidt, Florian},
  journal={arXiv preprint arXiv:1910.00292},
  year={2019}
}

@article{brown2024large,
  title={Large language monkeys: Scaling inference compute with repeated sampling},
  author={Brown, Bradley and Juravsky, Jordan and Ehrlich, Ryan and Clark, Ronald and Le, Quoc V and R{\'e}, Christopher and Mirhoseini, Azalia},
  journal={arXiv preprint arXiv:2407.21787},
  year={2024}
}

@article{snell2024scaling,
  title={Scaling llm test-time compute optimally can be more effective than scaling model parameters},
  author={Snell, Charlie and Lee, Jaehoon and Xu, Kelvin and Kumar, Aviral},
  journal={arXiv preprint arXiv:2408.03314},
  year={2024}
}

@article{bon,
  title={Inference-time scaling for diffusion models beyond scaling denoising steps},
  author={Ma, Nanye and Tong, Shangyuan and Jia, Haolin and Hu, Hexiang and Su, Yu-Chuan and Zhang, Mingda and Yang, Xuan and Li, Yandong and Jaakkola, Tommi and Jia, Xuhui and others},
  journal={arXiv preprint arXiv:2501.09732},
  year={2025}
}

@article{oshima2025inference,
  title={Inference-time text-to-video alignment with diffusion latent beam search},
  author={Oshima, Yuta and Suzuki, Masahiro and Matsuo, Yutaka and Furuta, Hiroki},
  journal={arXiv preprint arXiv:2501.19252},
  year={2025}
}

@article{scalingnoise,
  title={Scalingnoise: Scaling inference-time search for generating infinite videos},
  author={Yang, Haolin and Tang, Feilong and Hu, Ming and Yin, Qingyu and Li, Yulong and Liu, Yexin and Peng, Zelin and Gao, Peng and He, Junjun and Ge, Zongyuan and others},
  journal={arXiv preprint arXiv:2503.16400},
  year={2025}
}

@article{karthik2023if,
  title={If at first you don't succeed, try, try again: Faithful diffusion-based text-to-image generation by selection},
  author={Karthik, Shyamgopal and Roth, Karsten and Mancini, Massimiliano and Akata, Zeynep},
  journal={arXiv preprint arXiv:2305.13308},
  year={2023}
}

@article{qi2024not,
  title={Not all noises are created equally: Diffusion noise selection and optimization},
  author={Qi, Zipeng and Bai, Lichen and Xiong, Haoyi and Xie, Zeke},
  journal={arXiv preprint arXiv:2407.14041},
  year={2024}
}

@article{titan,
  title={TITAN-Guide: Taming Inference-Time AligNment for Guided Text-to-Video Diffusion Models},
  author={Simon, Christian and Ishii, Masato and Hayakawa, Akio and Zhong, Zhi and Takahashi, Shusuke and Shibuya, Takashi and Mitsufuji, Yuki},
  journal={arXiv preprint arXiv:2508.00289},
  year={2025}
}

@article{reno,
  title={Reno: Enhancing one-step text-to-image models through reward-based noise optimization},
  author={Eyring, Luca and Karthik, Shyamgopal and Roth, Karsten and Dosovitskiy, Alexey and Akata, Zeynep},
  journal={Advances in Neural Information Processing Systems},
  volume={37},
  pages={125487--125519},
  year={2024}
}

@article{pickscore,
  title={Pick-a-pic: An open dataset of user preferences for text-to-image generation},
  author={Kirstain, Yuval and Polyak, Adam and Singer, Uriel and Matiana, Shahbuland and Penna, Joe and Levy, Omer},
  journal={Advances in neural information processing systems},
  volume={36},
  pages={36652--36663},
  year={2023}
}

@article{hpsv2,
  title={Human preference score v2: A solid benchmark for evaluating human preferences of text-to-image synthesis},
  author={Wu, Xiaoshi and Hao, Yiming and Sun, Keqiang and Chen, Yixiong and Zhu, Feng and Zhao, Rui and Li, Hongsheng},
  journal={arXiv preprint arXiv:2306.09341},
  year={2023}
}

@article{imagereward,
  title={Imagereward: Learning and evaluating human preferences for text-to-image generation},
  author={Xu, Jiazheng and Liu, Xiao and Wu, Yuchen and Tong, Yuxuan and Li, Qinkai and Ding, Ming and Tang, Jie and Dong, Yuxiao},
  journal={Advances in Neural Information Processing Systems},
  volume={36},
  pages={15903--15935},
  year={2023}
}

@article{internvid,
  title={Internvid: A large-scale video-text dataset for multimodal understanding and generation},
  author={Wang, Yi and He, Yinan and Li, Yizhuo and Li, Kunchang and Yu, Jiashuo and Ma, Xin and Li, Xinhao and Chen, Guo and Chen, Xinyuan and Wang, Yaohui and others},
  journal={arXiv preprint arXiv:2307.06942},
  year={2023}
}

@article{goldennoise,
  title={Golden noise for diffusion models: A learning framework},
  author={Zhou, Zikai and Shao, Shitong and Bai, Lichen and Zhang, Shufei and Xu, Zhiqiang and Han, Bo and Xie, Zeke},
  journal={arXiv preprint arXiv:2411.09502},
  year={2024}
}

@article{ahn2024noise,
  title={A noise is worth diffusion guidance},
  author={Ahn, Donghoon and Kang, Jiwon and Lee, Sanghyun and Min, Jaewon and Kim, Minjae and Jang, Wooseok and Cho, Hyoungwon and Paul, Sayak and Kim, SeonHwa and Cha, Eunju and others},
  journal={arXiv preprint arXiv:2412.03895},
  year={2024}
}

@article{hypernoise,
  title={Noise Hypernetworks: Amortizing Test-Time Compute in Diffusion Models},
  author={Eyring, Luca and Karthik, Shyamgopal and Dosovitskiy, Alexey and Ruiz, Nataniel and Akata, Zeynep},
  journal={arXiv preprint arXiv:2508.09968},
  year={2025}
}

@article{lipman2022flow,
  title={Flow matching for generative modeling},
  author={Lipman, Yaron and Chen, Ricky TQ and Ben-Hamu, Heli and Nickel, Maximilian and Le, Matt},
  journal={arXiv preprint arXiv:2210.02747},
  year={2022}
}

@article{teacherforcing,
  title={Professor forcing: A new algorithm for training recurrent networks},
  author={Lamb, Alex M and ALIAS PARTH GOYAL, Anirudh Goyal and Zhang, Ying and Zhang, Saizheng and Courville, Aaron C and Bengio, Yoshua},
  journal={Advances in neural information processing systems},
  volume={29},
  year={2016}
}

@article{rollingforcing,
  title={Rolling Forcing: Autoregressive Long Video Diffusion in Real Time},
  author={Liu, Kunhao and Hu, Wenbo and Xu, Jiale and Shan, Ying and Lu, Shijian},
  journal={arXiv preprint arXiv:2509.25161},
  year={2025}
}

@article{selfforcing++,
  title={Self-Forcing++: Towards Minute-Scale High-Quality Video Generation},
  author={Cui, Justin and Wu, Jie and Li, Ming and Yang, Tao and Li, Xiaojie and Wang, Rui and Bai, Andrew and Ban, Yuanhao and Hsieh, Cho-Jui},
  journal={arXiv preprint arXiv:2510.02283},
  year={2025}
}

@article{anderson1982reverse,
  title={Reverse-time diffusion equation models},
  author={Anderson, Brian DO},
  journal={Stochastic Processes and their Applications},
  volume={12},
  number={3},
  pages={313--326},
  year={1982},
  publisher={Elsevier}
}

@article{consistencymodels,
  title={Consistency models},
  author={Song, Yang and Dhariwal, Prafulla and Chen, Mark and Sutskever, Ilya},
  year={2023}
}

@inproceedings{burgert2025go,
  title={Go-with-the-flow: Motion-controllable video diffusion models using real-time warped noise},
  author={Burgert, Ryan and Xu, Yuancheng and Xian, Wenqi and Pilarski, Oliver and Clausen, Pascal and He, Mingming and Ma, Li and Deng, Yitong and Li, Lingxiao and Mousavi, Mohsen and others},
  booktitle={Proceedings of the Computer Vision and Pattern Recognition Conference},
  pages={13--23},
  year={2025}
}

@article{howiwarpyournoise,
  title={How i warped your noise: a temporally-correlated noise prior for diffusion models},
  author={Chang, Pascal and Tang, Jingwei and Gross, Markus and Azevedo, Vinicius C},
  journal={arXiv preprint arXiv:2504.03072},
  year={2025}
}

@article{xiao2025worldmem,
  title={Worldmem: Long-term consistent world simulation with memory},
  author={Xiao, Zeqi and Lan, Yushi and Zhou, Yifan and Ouyang, Wenqi and Yang, Shuai and Zeng, Yanhong and Pan, Xingang},
  journal={arXiv preprint arXiv:2504.12369},
  year={2025}
}

@article{po2025long,
  title={Long-context state-space video world models},
  author={Po, Ryan and Nitzan, Yotam and Zhang, Richard and Chen, Berlin and Dao, Tri and Shechtman, Eli and Wetzstein, Gordon and Huang, Xun},
  journal={arXiv preprint arXiv:2505.20171},
  year={2025}
}

@article{cheng2025playing,
  title={Playing with Transformer at 30+ FPS via Next-Frame Diffusion},
  author={Cheng, Xinle and He, Tianyu and Xu, Jiayi and Guo, Junliang and He, Di and Bian, Jiang},
  journal={arXiv preprint arXiv:2506.01380},
  year={2025}
}

@misc{oasis2024,
  title        = {Oasis: A Universe in a Transformer},
  author       = {Decart, Julian and Quevedo, Julian and McIntyre, Quinn and Campbell, Spruce and Chen, Xinlei and Wachen, Robert},
  year         = {2024},
  howpublished = {\url{https://oasis-model.github.io/}}
}

@article{Captaincinema,
  title={Captain cinema: Towards short movie generation},
  author={Xiao, Junfei and Yang, Ceyuan and Zhang, Lvmin and Cai, Shengqu and Zhao, Yang and Guo, Yuwei and Wetzstein, Gordon and Agrawala, Maneesh and Yuille, Alan and Jiang, Lu},
  journal={arXiv preprint arXiv:2507.18634},
  year={2025}
}

@article{moviedreamer,
  title={Moviedreamer: Hierarchical generation for coherent long visual sequence},
  author={Zhao, Canyu and Liu, Mingyu and Wang, Wen and Chen, Weihua and Wang, Fan and Chen, Hao and Zhang, Bo and Shen, Chunhua},
  journal={arXiv preprint arXiv:2407.16655},
  year={2024}
}

@misc{D-Flow,
      title={D-Flow: Differentiating through Flows for Controlled Generation}, 
      author={Heli Ben-Hamu and Omri Puny and Itai Gat and Brian Karrer and Uriel Singer and Yaron Lipman},
      year={2024},
      eprint={2402.14017},
      archivePrefix={arXiv},
      primaryClass={cs.LG},
      url={https://arxiv.org/abs/2402.14017}, 
}

@misc{guo2025trainingfreeguidancedifferentiabilityscalable,
      title={Training-Free Guidance Beyond Differentiability: Scalable Path Steering with Tree Search in Diffusion and Flow Models}, 
      author={Yingqing Guo and Yukang Yang and Hui Yuan and Mengdi Wang},
      year={2025},
      eprint={2502.11420},
      archivePrefix={arXiv},
      primaryClass={cs.LG},
      url={https://arxiv.org/abs/2502.11420}, 
}

@article{li2024t2v,
  title={T2v-turbo-v2: Enhancing video generation model post-training through data, reward, and conditional guidance design},
  author={Li, Jiachen and Long, Qian and Zheng, Jian and Gao, Xiaofeng and Piramuthu, Robinson and Chen, Wenhu and Wang, William Yang},
  journal={arXiv preprint arXiv:2410.05677},
  year={2024}
}

@inproceedings{internvideo2,
  title={Internvideo2: Scaling foundation models for multimodal video understanding},
  author={Wang, Yi and Li, Kunchang and Li, Xinhao and Yu, Jiashuo and He, Yinan and Chen, Guo and Pei, Baoqi and Zheng, Rongkun and Wang, Zun and Shi, Yansong and others},
  booktitle={European Conference on Computer Vision},
  pages={396--416},
  year={2024},
  organization={Springer}
}

@article{vidprom,
  title={Vidprom: A million-scale real prompt-gallery dataset for text-to-video diffusion models},
  author={Wang, Wenhao and Yang, Yi},
  journal={Advances in Neural Information Processing Systems},
  volume={37},
  pages={65618--65642},
  year={2024}
}

@article{tag2text,
  title={Tag2text: Guiding vision-language model via image tagging},
  author={Huang, Xinyu and Zhang, Youcai and Ma, Jinyu and Tian, Weiwei and Feng, Rui and Zhang, Yuejie and Li, Yaqian and Guo, Yandong and Zhang, Lei},
  journal={arXiv preprint arXiv:2303.05657},
  year={2023}
}

@article{Selfreflection,
  title={Self-reflection in llm agents: Effects on problem-solving performance},
  author={Renze, Matthew and Guven, Erhan},
  journal={arXiv preprint arXiv:2405.06682},
  year={2024}
}

@article{rope,
  title={Roformer: Enhanced transformer with rotary position embedding},
  author={Su, Jianlin and Ahmed, Murtadha and Lu, Yu and Pan, Shengfeng and Bo, Wen and Liu, Yunfeng},
  journal={Neurocomputing},
  volume={568},
  pages={127063},
  year={2024},
  publisher={Elsevier}
}

@article{longlive,
  title={Longlive: Real-time interactive long video generation},
  author={Yang, Shuai and Huang, Wei and Chu, Ruihang and Xiao, Yicheng and Zhao, Yuyang and Wang, Xianbang and Li, Muyang and Xie, Enze and Chen, Yingcong and Lu, Yao and others},
  journal={arXiv preprint arXiv:2509.22622},
  year={2025}
}

@article{uehara2025inference,
  title={Inference-time alignment in diffusion models with reward-guided generation: Tutorial and review},
  author={Uehara, Masatoshi and Zhao, Yulai and Wang, Chenyu and Li, Xiner and Regev, Aviv and Levine, Sergey and Biancalani, Tommaso},
  journal={arXiv preprint arXiv:2501.09685},
  year={2025}
}

@article{adamw,
  title={Decoupled weight decay regularization},
  author={Loshchilov, Ilya and Hutter, Frank},
  journal={arXiv preprint arXiv:1711.05101},
  year={2017}
}

@article{tweedie1,
  title={Tweedie’s formula and selection bias},
  author={Efron, Bradley},
  journal={Journal of the American Statistical Association},
  volume={106},
  number={496},
  pages={1602--1614},
  year={2011},
  publisher={Taylor \& Francis}
}

@incollection{tweedie2,
  title={An empirical Bayes approach to statistics},
  author={Robbins, Herbert E},
  booktitle={Breakthroughs in Statistics: Foundations and basic theory},
  pages={388--394},
  year={1992},
  publisher={Springer}
}

@article{ctm,
  title={Consistency trajectory models: Learning probability flow ode trajectory of diffusion},
  author={Kim, Dongjun and Lai, Chieh-Hsin and Liao, Wei-Hsiang and Murata, Naoki and Takida, Yuhta and Uesaka, Toshimitsu and He, Yutong and Mitsufuji, Yuki and Ermon, Stefano},
  journal={arXiv preprint arXiv:2310.02279},
  year={2023}
}

@article{dinov2,
  title={Dinov2: Learning robust visual features without supervision},
  author={Oquab, Maxime and Darcet, Timoth{\'e}e and Moutakanni, Th{\'e}o and Vo, Huy and Szafraniec, Marc and Khalidov, Vasil and Fernandez, Pierre and Haziza, Daniel and Massa, Francisco and El-Nouby, Alaaeldin and others},
  journal={arXiv preprint arXiv:2304.07193},
  year={2023}
}
}
\clearpage
\newpage
\clearpage
\setcounter{section}{0}
\renewcommand{\thesection}{\Alph{section}}
\renewcommand{\thesubsection}{\thesection.\arabic{subsection}}
\renewcommand{\thesubsubsection}{\thesubsection.\arabic{subsubsection}}
\setcounter{page}{1}
\maketitlesupplementary
\section{Details on DMD Loss and Samplers}
We provide additional details on the formulation of the step-distillation objective using DMD loss, along with descriptions of the corresponding stochastic and deterministic sampling procedures discussed in \cref{sec:traj_noise_refine} of the main paper.
\paragraph{Detailed Formulation of DMD Loss.} Following previous works~\cite{wang2023prolificdreamer,dmd}, the gradient of reverse KL divergence in \cref{eq:dmd} of the main paper is obtained by 
\begin{equation}
\begin{aligned}
& \nabla_\theta \mathcal{L}_{\mathrm{DMD}} \\
& =  \mathbb{E}_t\left(\nabla_\theta \operatorname{KL}\left(q_{\theta , t}\left(\bx_t\right) \| p_{t}\left(\bx_t\right)\right)\right) \\
& = \mathbb{E}_t\left(\nabla_\theta \left(\mathbb{E}_{\bx_t \sim p_{t|0}\left(\bx_t|\bx\right),\bx \sim q_{\theta}\left(\bx\right)}\log \left(\frac{q_{\theta , t}\left(\bx_t\right)}{p_{t}\left(\bx_t\right)}\right)\right) \right) \\
& = \mathbb{E}_{t,\bx_t \sim p_{t|0}\left(\bx_t|\bx\right),\bx \sim q_{\theta}\left(\bx\right)}\left[s_{\text{fake}} \left( \bx_t, t\right)-s_{\text{real}} \left( \bx_t, t\right) \right]\frac{d\bx_t}{d\theta},
\end{aligned}
\label{eq:supp_dmd}
\end{equation}
where sampling from the conditional $p_{t|0}$ corresponds to the forward diffusion process in \cref{eq:forward_diff} of the main paper. In practice, the samples from the model distribution $q_{\theta}\left(\bx\right)$ are obtained autoregressively as 
\begin{equation}
q_\theta\left(\bx_t\right)=q_\theta\left(\bx_t^{1: N}\right)=\prod_{i=1}^N q_\theta\left(\bx_t^i \mid \bx_t^{<i}\right).
\end{equation}
$s_{\text{fake}}$ and $s_{\text{real}}$ are score functions for the model distribution and real-data distribution, \ie, 
\begin{equation}
    \begin{aligned}
        s_{\text{fake}} \left( \bx_t, t\right)
        & =\nabla_{\bx_t}\log q_\theta\left(\bx_t\right),\\
        s_{\text{real}} \left( \bx_t, t\right)&=\nabla_{\bx_t}\log p\left(\bx_t\right).
    \end{aligned}
\label{eq:scores}
\end{equation}
 
\paragraph{Few-step stochastic sampling.}
Both base AR-VDMs used in our method---Self-Forcing and CausVid are step-distilled from a standard multi-step bidirectional video diffusion model~\cite{wan2025wan} that follows the training objective of Rectified Flows~\cite{lipman2022flow,liu2022flow}, with which the forward diffusion schedule is defined as
\begin{equation}
    \bx_t = t\bx_0 + \left(1-t\right)\bx_{T_\mathrm{max}}.
\end{equation}
This constitutes a ground truth velocity field of 
\begin{equation}
    \bv_t=\frac{d \bx_t}{d t}=\bx_0-\bx_{T_\mathrm{max}},
\end{equation}
with which the training objective is defined as mean squared error (MSE) between ground truth velocity and predicted velocity from a flow-based model $\bv_{\theta_0}$
\begin{equation}
\mathcal{L}_{\text{flow}}=\mathbb{E}_{\bx_0, \bx_{T_\mathrm{max}}, t}\left\|\bv_{\theta_0}\left(\bx_t, t\right)-\bv_t\right\|^2.
\end{equation}
Fine-tuned based on the model $\bv_{\theta_0}$, a denoising prediction step of Self-Forcing and CausVid in \cref{eq:denoising_step} of the main paper is parameterized in a form similar to Tweedie's estimation~\cite{tweedie1,tweedie2} 
\begin{equation}
\begin{aligned}
{\bx}^i_{0|{t_{j}}}&=G_\theta\left(\bx_{t_j}^i;\bx^{<i},t_j\right)\\
&=\bx_{t_j}^i+\left(1-t_j\right)\bv_\theta\left(\bx_{t_j}^i;\bx^{<i},t_j\right),
\end{aligned}
\label{eq:supp_denoising}
\end{equation}
where $\bv_\theta$ shares the same architecture as $\bv_{\theta_0}$. $\bv_\theta$ replaces bidirectional attention masks with causal attention masks in self-attention layers, which in turn enables the use of KV cache to condition on historical frames $\bx^{<i}$.
\paragraph{Implementation of ODE-based sampling.} For the example of ODE sampling in \cref{fig:ode_vs_stoch} of the main paper. We change a stochastic sampling step of 
\begin{equation}
\begin{aligned}
f_{\theta, t_j}\left(\bx_{t_j}^i\right)=\Psi\left({\bx}_{0|t_j}^i,\epsilon^i_{j-1}, t_{j-1}\right),
\end{aligned}
\end{equation}to the following
\begin{equation}
\begin{aligned}
\tilde{f}_{\theta, t_j}\left(\bx_{t_j}^i\right)=\bx_{t_j}^i+\left(t_{j-1}-t_j\right)\bv_\theta\left(\bx_{t_j}^i;\bx^{<i},t_j\right),
\end{aligned}
\end{equation}
which is the same ODE sampling schedule used by the base bidirectional model $\bv_{\theta_0}$.
 \begin{table*}[!ht]
\centering
\resizebox{\linewidth}{!}{
\footnotesize
\begin{tabular}{l c c | c c c c c c c}
    \toprule
    & \makecell[c]{\textbf{Total}\\ \textbf{Score} $\uparrow$} & \makecell[c]{\textbf{Quality}\\ \textbf{Score} $\uparrow$}
    & \makecell[c]{Subject\\Consist. $\uparrow$}
    & \makecell[c]{Background\\Consist. $\uparrow$}
    & \makecell[c]{Aesthetic\\Score $\uparrow$}
    & \makecell[c]{Imaging\\Quality $\uparrow$}
    & \makecell[c]{Temporal\\Flicker $\uparrow$}
    & \makecell[c]{Motion\\Smooth. $\uparrow$}
    & \makecell[c]{Dynamic\\Degree $\uparrow$} \\
    \midrule
    \rowcolor{lightgray}Self-Forcing~\cite{selfforcing} & 84.05 & 84.93 & 95.02 & 96.39 & 65.78 & 69.24 & 99.11 & 98.42 & 69.44 \\
    +LoRA & 82.10 & 82.65 & 93.17 & 92.96 & 63.90 & 66.64 & 95.43 & 96.42 & 95.83 \\
    +InitNoiseRefiner & 83.76 & 84.63 & 94.41 & 95.79 & 65.19 & 68.78 & 99.01 & 98.26 & 72.22 \\
    \rowcolor{lightblue}+AutoRefiner (ours) & \textbf{84.72} & \textbf{85.41} & 95.42 & 96.17 & 65.84 & 69.33 & 99.09 & 98.24 & 76.39 \\
    \midrule 
    \rowcolor{lightgray}CausVid~\cite{causvid} & 83.06 & 84.24 & 96.24 & 95.83 & 64.71 & 68.33 & 99.41 & 98.07 & 63.89 \\
    +LoRA & 83.43 & 84.74 & 94.97 & 95.14 & 62.50 & 71.65 & 97.48 & 97.27 & 88.89\\
    +InitNoiseRefiner & 82.23 & 83.13 & 96.50 & 96.74 & 65.18 & 67.68 & 99.53 & 98.31 & 44.44 \\
    \rowcolor{lightblue}+AutoRefiner (ours) & \textbf{83.84} & \textbf{84.82} & 95.55 & 95.86 & 65.13 & 69.10 & 99.25 & 98.36 & 69.44\\
    \bottomrule
\end{tabular}}
\resizebox{\linewidth}{!}{
\footnotesize
\begin{tabular}{l c | c c c c c c c c c}
    \toprule 
    & \makecell[c]{\textbf{Semantic}\\ \textbf{Score} $\uparrow$}
    & \makecell[c]{Object\\Class $\uparrow$}
    & \makecell[c]{Multiple\\Objects $\uparrow$}
    & \makecell[c]{Color$\uparrow$}
    & \makecell[c]{Spatial\\Relation $\uparrow$}
    & \makecell[c]{Scene $\uparrow$}
    & \makecell[c]{Temporal\\Style $\uparrow$}
    & \makecell[c]{Overall\\Consistency $\uparrow$}
    & \makecell[c]{Human\\Action $\uparrow$} & \makecell[c]{Appearance\\Style $\uparrow$}\\
    \midrule
    \rowcolor{lightgray}Self-Forcing~\cite{selfforcing} & 80.54 & 94.22 & 88.41 & 86.81 & 81.24 & 52.83 & 24.58 & 26.80 & 97 & 20.52\\
    +LoRA & 79.93 & 90.90 & 83.08 & 88.11 & 78.92 & 56.32 & 24.93 & 27.08 & 96 & 20.27 \\
    +InitNoiseRefiner & 80.29 & 92.17 & 86.51 & 85.72 & 82.71 & 53.78 & 24.49 & 27.05 & 97 & 20.45 \\
    \rowcolor{lightblue}+AutoRefiner (ours) & \textbf{81.95} & 93.83 & 88.64 & 88.99 & 86.56 & 57.20 & 24.59 & 26.71 & 98 & 20.31 \\
    \midrule 
    \rowcolor{lightgray}CausVid~\cite{causvid}& 78.35 & 94.22 & 88.03 & 86.46 & 74.04 & 48.47 & 23.93 & 25.77 & 96 & 20.25 \\
    +LoRA & 78.18 & 89.95 & 82.62 & 86.00 & 73.32 & 58.79 & 24.31 & 25.93 & 91 & 20.36\\
    +InitNoiseRefiner & 78.64 & 93.83 & 87.27 & 88.16 & 76.44 & 46.95 & 23.72 & 25.68 & 97 & 20.65 \\
    \rowcolor{lightblue}+AutoRefiner (ours) & \textbf{79.93} & 95.25 & 86.43 & 89.78 & 79.30 & 53.05 & 23.91 & 26.21 & 97 & 19.84\\
    \bottomrule
\end{tabular}}
\caption{VBench results across all 16 evaluation dimensions for comparison with baseline methods using DMD loss as objective.}
\label{tab:supp_vbench_metrics}
\end{table*}
\section{Implementation Details}
We provide more implementation details on our used base AR-VDMs and training configurations.
\paragraph{Base AR-VDMs.} Our used base AR-VDMs---Self-Forcing and CausVid are both distilled from the bidirectional 1.3B Wan2.1~\cite{wan2025wan} model. Self-Forcing is distilled to generate a video chunk using four denoising steps $\{1000, 750, 500, 250\}$.
CausVid is distilled to generate a video chunk using three denoising steps $\{1000, 757, 522\}$. Both models generate a total of 21 latent frame videos with 3-frame chunks generated at each autoregressive rollout, which are decoded into 81-frame videos at a spatial resolution of $480\times832$ by a 3D VAE~\cite{wan2025wan}. 
\paragraph{Training Details.} For training with DMD loss, we use an AdamW~\cite{adamw} optimizer with a learning rate $1\times10^{-4}$. For the score functions in \cref{eq:scores}, we adopt a Wan2.1-14B model as the real score function and a Wan2.1-1.3B model as the fake score function. For training with reward scores, we use an AdamW optimizer with a learning rate of $2\times10^{-5}$. The reward weights are set to $0.25$, $0.1$ and $0.1$ for InternVideo2, HPSv2.1 and CLIP, respectively. Among the 81 generated frames, we randomly sample 2 frames to compute image reward scores, and extract 4-frame clips with a sampling interval of 10 frames to compute video reward scores. Following the configurations of the base models, we use a classifier-free guidance (CFG)~\cite{classifier_free_guidance} scale of $3.0$ for Self-Forcing and $3.5$ for CausVid. The timestep shift~\cite{sd3} is set to $5.0$ for Self-Forcing and $8.0$ for CausVid. We apply $L_2$ regularization for reward-based training and not for DMD loss-based training, as the reverse-KL formulation of the DMD loss in \cref{eq:supp_dmd} inherently regularizes the noise refiner by keeping its tilted distribution close to the base model distribution. We set both LoRA rank and alpha as 128. All trainings are performed on 8 H100 GPUs with a valid batch size of 48 for 1,000 iterations, taking roughly 24 hours to complete. We use fully shared data parallel (FSDP) and gradient checkpointing to reduce the maximum GPU memory usage.
\vspace{-0.2cm}
\section{Further Quantitative Results}
We provide additional quantitative evaluations, including full VBench scores across all dimensions, an ablation study on the LoRA rank, and a comparison of how different training-based methods influence the sample diversity and dynamic degree of the base AR-VDM.
\paragraph{Full VBench Scores.} We provide in \cref{tab:supp_vbench_metrics} the full VBench scores across all 16 dimensions corresponding to the main comparison results reported in \cref{tab:dmd_comparison} of the main paper, including 7 quality-related metrics and 9 semantic-related metrics.
\paragraph{Ablation On LoRA rank.} We report the ablation results for different LoRA ranks in \cref{tab:ablation_rank}.
\begin{table}[!h]
\centering
\resizebox{\linewidth}{!}{
\footnotesize
\begin{tabular}{cccccc}
    \toprule  
    LoRA Rank & {Quality Score $\uparrow$} & {Semantic Score $\uparrow$} & {Total Score $\uparrow$} \\
    \midrule
    16 & 84.97 & \textbf{82.00} & 84.38 \\
    32 & 84.72 & 80.93 & 83.97 \\
    64 & 84.93 & 81.52 & 84.25 \\
    128 & \textbf{85.41} & 81.95 & \textbf{84.72} \\
    \bottomrule
\end{tabular}}
\caption{Ablation on different LoRA ranks.}
\label{tab:ablation_rank}
\end{table}
\vspace{-0.5cm}
\paragraph{Diversity \& Dynamic Preservation.} We further analyze how reward-based training with different methods impacts the sample diversity and dynamic degree of the base AR-VDM using image similarity metrics. For each method, we generate 10 videos using different random seeds for each of 70 text prompts. Sample diversity is computed as the average cosine similarity between the embeddings of all video pairs generated from the same prompt, where each video embedding is obtained by averaging its frame-level DINO~\cite{dinov2} embedding. Dynamic degree is evaluated by the similarity between temporally neighboring frames within each video using LPIPS~
\cite{lpips}, SSIM~\cite{SSIM} and PSNR~\cite{PSNR}, with a sampling interval of 5 frames. The results are reported in \cref{tab:supp_diversity}. We observe that, while LoRA exhibits mode-collapse behavior reflected by significantly increased DINO similarity, \name preserves a sample diversity comparable to the base model. Moreover, both LoRA and InitNoiseRefiner substantially degrade the dynamic degree---showing increased similarity in LPIPS, SSIM, and PSNR scores relative to the base model---whereas \name consistently improves these metrics. 
\begin{table}[h]
\centering
\resizebox{\linewidth}{!}{
\footnotesize
\begin{tabular}{l c | ccc}
    \toprule  
    \multirow{2}{*}{Method} & \multicolumn{1}{c|}{{Diversity}} 
    & \multicolumn{3}{c}{{Dynamic Degree}} \\
    \cmidrule(lr){2-5}   
    & DINO$\downarrow$ & LPIPS$\uparrow$ & SSIM$\downarrow$ & PSNR$\downarrow$ \\
    \midrule
    \rowcolor{lightgray}Self-Forcing~\cite{selfforcing} & .8577 & .1359 & .5636 & 17.52 \\
    +InitNoiseRefiner & .8394 & .0652 & .6807 & 19.47 \\
    +LoRA & .9039 & .0488 & .7430 & 19.32 \\
    \rowcolor{lightblue}+AutoRefiner (ours) & .8597 & .1758 & .4393 & 14.25 \\
    \bottomrule
\end{tabular}
}
\caption{Comparison of sample diversity and dynamic degree using frame-wise similarity metrics after reward-based training.}
\label{tab:supp_diversity}
\end{table}
\vspace{-.42cm}
\section{Further Qualitative Results}
\begin{figure}[h]    \centering\includegraphics[width=\linewidth]{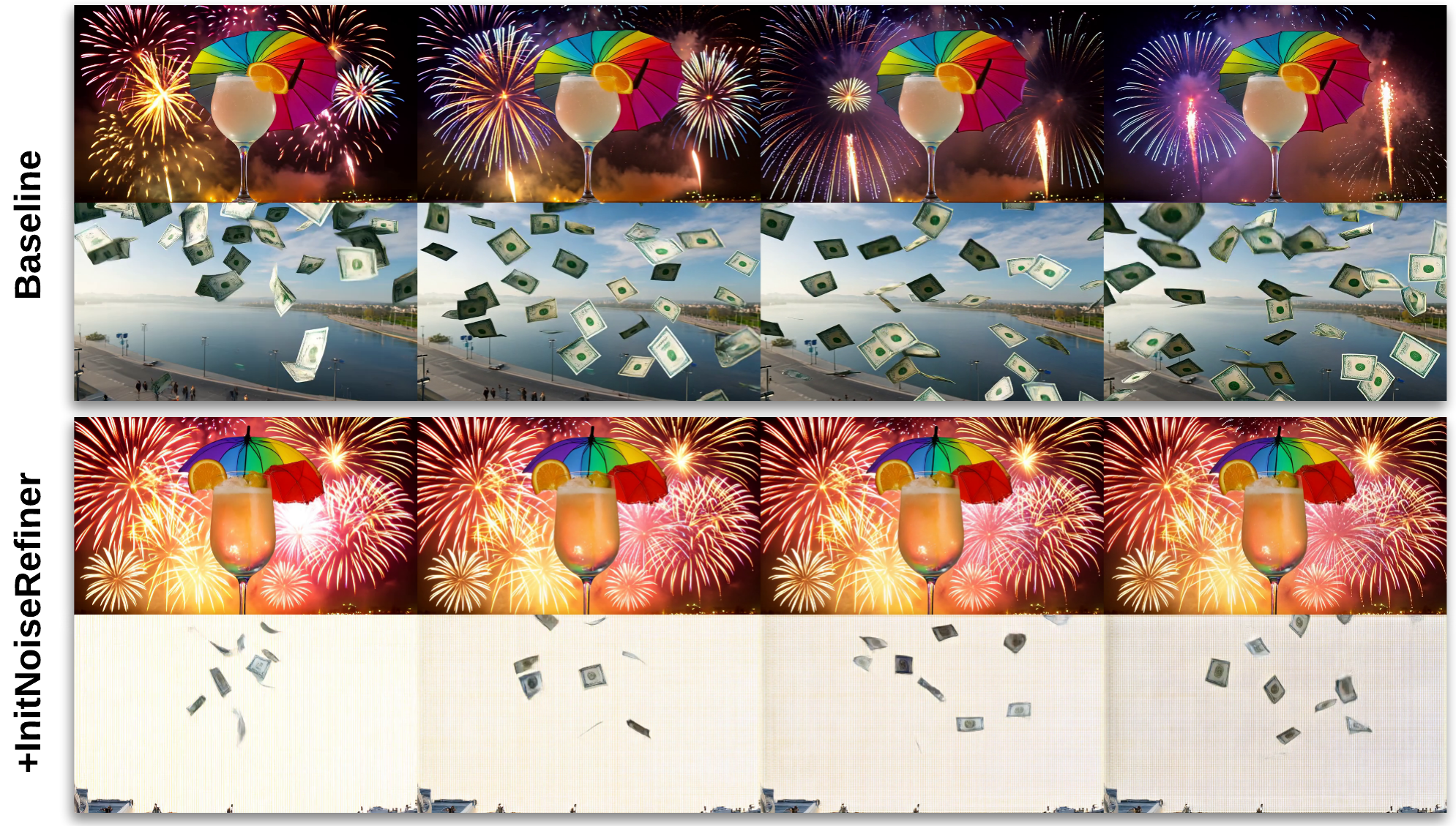}
    \caption{An illustration of two reward-hacking issues arising in initial noise refinement: static motion (as in the ``\textit{firework}" example) and ``grid-like" artifacts (as in the ``\textit{money}" example).}
    \label{fig:init_refine_failure}
\end{figure}

\begin{figure*}[!ht]
    \centering
        \includegraphics[width=\linewidth]{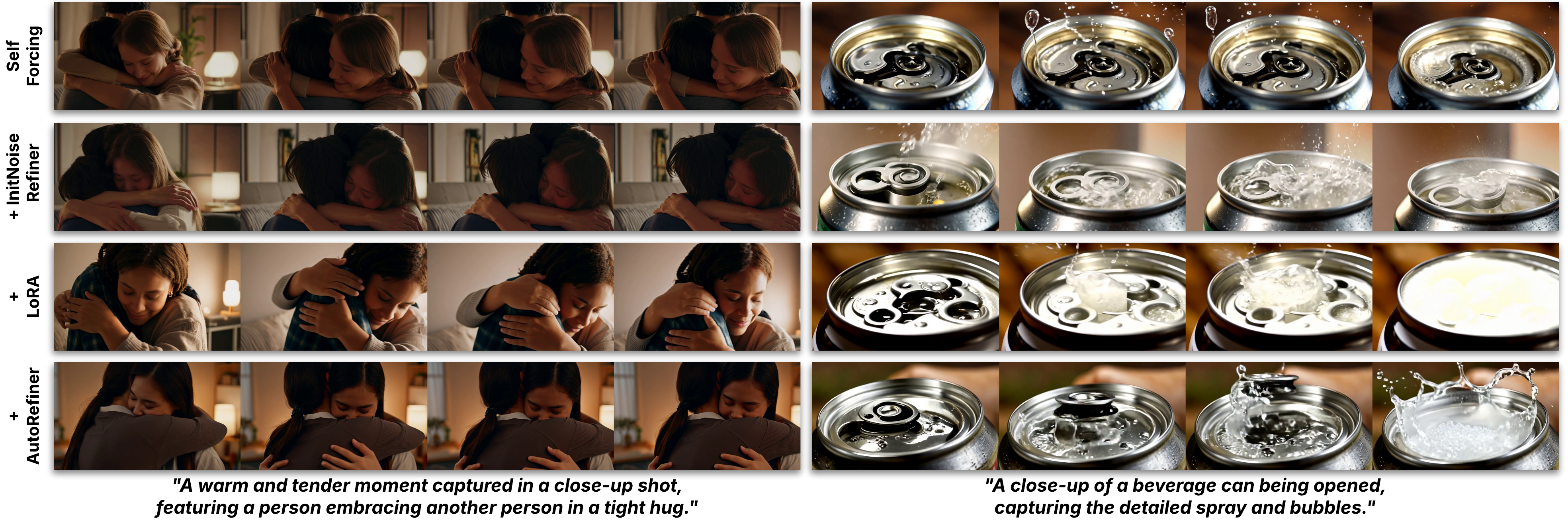}
        \caption{More qualitative comparison with baselines on training with DMD loss, using Self-Forcing~\cite{selfforcing} as the base model. \name alleviates visual artifacts (left example), and improves motion alignment with text prompt (right example) compared to the base model.} 
        \label{fig:dmd_sf_comparison_supp}
\end{figure*}

\begin{figure*}[ht]
    \centering
        \includegraphics[width=\linewidth]{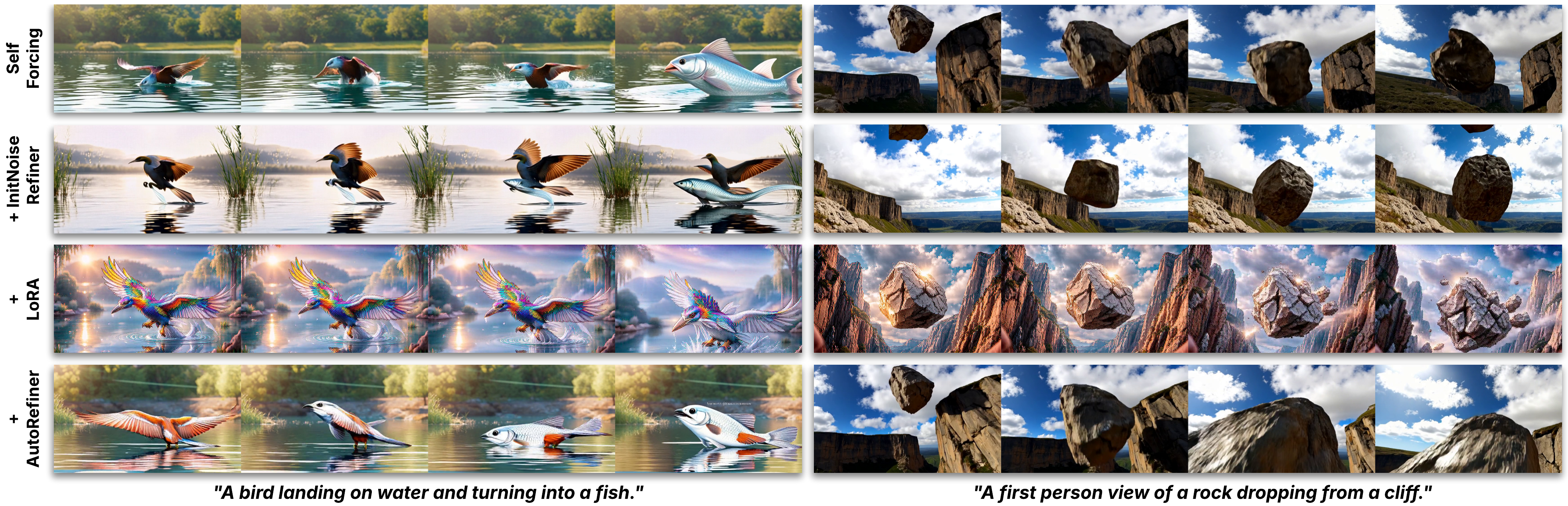}
        \caption{More qualitative comparison with baselines on training with reward scores. \name demonstrates smoother temporal transitions (left example) and generates motion that appears more natural and dynamic than the baselines (right example).}
        \label{fig:reward_comparison_supp}
\end{figure*}

We present additional qualitative comparisons with baseline methods, including DMD-based training using Self-Forcing as the base model (\cref{fig:dmd_sf_comparison_supp}), DMD-based training using CausVid as the base model (\cref{fig:dmd_cv_comparison_supp}), and reward-based training using Self-Forcing as the base model (\cref{fig:reward_comparison_supp}). In \cref{fig:init_refine_failure}, we illustrate the two reward hacking behaviors of training an initial noise refiner for AR-VDMs---motion degradation and `grid-like' artifacts, as discussed in \cref{sec:traj_noise_refine} of the main paper. Please refer to the HTML file in the supplementary materials to review the videos for all examples presented in both the main paper and the appendix.
\section{Subjective Evaluation}
We conduct a subjective evaluation with 10 participants to compare \name with several baseline methods. Each method generates videos for 10 text prompts. Participants are asked to watch each video together with its corresponding prompt and rate the overall video quality on a 1–5 scale (where higher ratings indicate better quality). During evaluation, participants are instructed to consider the following criteria:
\vspace{0.4cm}
\begin{itemize}
\item \textbf{Visual Quality}: How clear, realistic, and artifact-free the generated video appears.
\item \textbf{Temporal Naturalness \& Consistency}: How smooth and coherent the motion is across frames.
\item \textbf{Semantic Alignment}: How well the video content corresponds to the meaning of the text prompt.
\end{itemize}
\vspace{0.4cm}
The results are reported in \cref{tab:user_study}.
\begin{table}[ht]
\centering
\resizebox{.75\linewidth}{!}{
\footnotesize
\begin{tabular}{l c c c c}
    \toprule
    & User preference score$\uparrow$\\
    \midrule
    \rowcolor{lightgray}Self-Forcing~\cite{selfforcing} & 2.23\\
    +InitNoiseRefiner & 2.31\\
    +LoRA & 2.25\\
    \rowcolor{lightblue}+AutoRefiner (ours)  & 3.69 \\
    \bottomrule
\end{tabular}}
\caption{Comparison with baseline methods with subjective evaluation on a 1-5 scale.}
\label{tab:user_study}
\end{table}
\vspace{-3mm}
\section{Future Work} 
\begin{figure*}[ht]
    \centering
        \includegraphics[width=\linewidth]{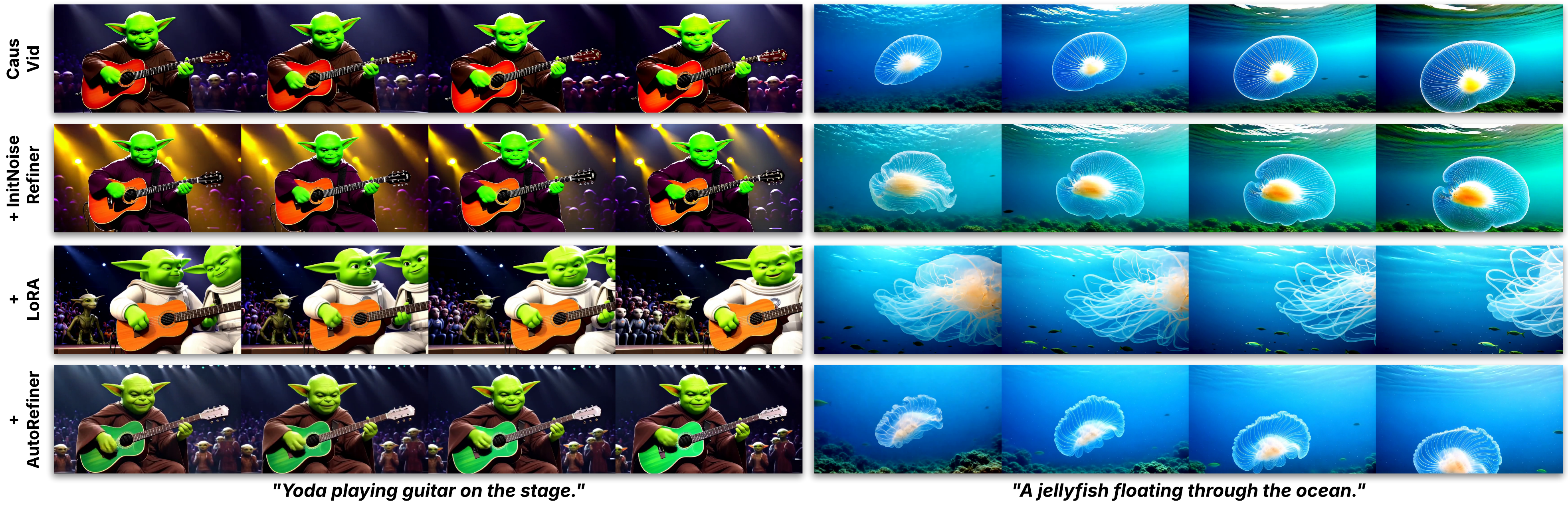}
        \caption{More qualitative comparison with baselines on training with DMD loss, using CausVid~\cite{causvid} as the base model. \name alleviates the cross-frame inconsistency issue of CausVid, as observed in the gradually saturated color of ``\textit{Yoda}" (left) and ``\textit{ocean}" (right).}
        \label{fig:dmd_cv_comparison_supp} 
\end{figure*}

Future work may extend the idea of noise refinement with self-reflection to further mitigate the exposure bias issue in frame extrapolation tasks, which is one of the core challenges in autoregressive generation for continuous data. We already observe such potential when applying \name to CausVid, as reflected by the reduced color-drifting artifacts in \cref{fig:dmd_cv_comparison} (main paper) and \cref{fig:dmd_cv_comparison_supp}. Another promising direction is to explore higher-level motion controllability with a noise refiner, such as controlling the speed or camera motion of generated videos, by steering the sampled noises. For image generation, future work may also investigate applying the concept of pathwise noise refinement to text-to-image models step-distilled by DMD or Consistency Distillation~\cite{consistencymodels,ctm}, enabling downstream improvements such as reward alignment or enhanced physical soundness.
\newpage


\end{document}